%% file: manuscript.tex
\documentclass[default,sn-mathphys]{sn-jnl}
\input{0-packages_commands}
\begin{document}
\input{1-acronyms}

\title[A ZSAR method based on natural language descriptions]{Tell me what you see: A zero-shot action recognition method based on natural language descriptions}

\author*[1,2]{\fnm{Valter} \sur{Estevam}}\email{valter.junior@ifpr.edu.br}
\author[2]{\fnm{Rayson} \sur{Laroca}}\email{rblsantos@inf.ufpr.br}
\author[3]{\fnm{Helio} \sur{Pedrini}}\email{helio@ic.unicamp.br}
\author[2]{\fnm{David} \sur{Menotti}}\email{menotti@inf.ufpr.br}

\affil[1]{\orgname{Federal Institute of Paran\'a}, \orgaddress{\city{Irati}, \postcode{84500-000}, \state{Paran\'a}, \country{Brazil}}}
\affil[2]{\orgdiv{Department of Informatics}, \orgname{Federal University of Paran\'a}, \orgaddress{\city{Curitiba}, \postcode{81531-970}, \state{Paran\'a}, \country{Brazil}}}
\affil[3]{\orgdiv{Institute of Computing},  \orgname{University of Campinas}, \orgaddress{\city{Campinas}, \postcode{13083-852}, \state{S\~ao Paulo}, \country{Brazil}}}

\abstract{
\major{This paper presents a novel approach to Zero-Shot Action Recognition.
Recent works have explored the detection and classification of objects to obtain semantic information from videos with remarkable performance.
Inspired by them, we propose using video captioning methods to extract semantic information about objects, scenes, humans, and their relationships.
To the best of our knowledge, this is the first work to represent both videos and labels with descriptive sentences.
More specifically, we represent videos using sentences generated via video captioning methods and classes using sentences extracted from documents acquired through search engines on the Internet. 
Using these representations, we build a shared semantic space employing BERT-based embedders pre-trained in the paraphrasing task on multiple text datasets.
The projection of both visual and semantic information onto this space is straightforward, as they are sentences, enabling classification using the nearest neighbor rule.
We demonstrate that representing videos and labels with sentences alleviates the domain adaptation problem.
Additionally, we show that word vectors are unsuitable for building the semantic embedding space of our descriptions.
Our method outperforms the state-of-the-art performance on the UCF101 dataset by 3.3 p.p. in accuracy under the TruZe protocol and achieves competitive results on both the UCF101 and HMDB51 datasets under the conventional protocol (0/50\% - training/testing split).
Our code is available at \url{https://github.com/valterlej/zsarcap}.}
}


\maketitle

{\let\thefootnote\relax\footnote{\copyrighttext}}

\section{Introduction}
\label{sec:introduction}

\ac{har} is an active research topic in computer vision. Several supervised models have been proposed with impressive performance in the last years, especially those based on deep learning~\cite{kong:2022survey}.
At the same time, large-scale datasets containing a massive number of human actions, such as Kinetics-400~\cite{carreira:2017}, Kinetics-700~\cite{carreira:2019} and ActivityNet~\cite{heilbron:2015}, have become available.
Even in the face of this progress, only a few human actions are mapped, collected and annotated.
Hence, retraining \ac{sota} action recognition models is imperative to incorporate new classes, which requires much time, computational resources, energy, and~human labor~\cite{wang:2017}.

\ac{zsl}~\cite{xie:2020,wang:2020} and their applications to actions,~\ac{zsar}~\cite{wang:2017,chen:2021,mettes:2021}, are computer vision tasks that emerge from this problem. 
In \ac{zsar}, the goal is to recognize examples from unknown human action classes, that is, videos from classes that were not available during the training stage. 
As we do not have samples from a new class in training, \ac{zsar} models need to represent the class labels with semantic information, and the classification is performed with some function, usually learned with known classes by correlating visual patterns with the label semantic~properties~\cite{estevam:2021}.

Traditionally, the videos are represented using spatio-temporal features (\eg \ac{idt}~\cite{wang:2013}, \ac{c3d}~\cite{tran:2015} or \ac{i3d}~\cite{carreira:2017}), and the class labels are represented with attributes or word vectors such as Word2Vec~\cite{mikolov:2013} or \ac{glove}~\cite{pennington:2014}. 

\major{Although this general scheme (deep features $ \leftrightarrow $ word vectors) has become popular for~\ac{zsar}, it suffers from a severe domain adaption problem because the learned functions do not transfer well from seen to unseen classes. The main reason is the gap between visual features and semantic features represented with word vectors~\cite{wang:2017}}. For example, different concepts such as \textit{horse riding} and \textit{pommel horse} are prone to appear close into the semantic space, and the absence of complementary information makes it very difficult to discriminate them.
It is not surprising that attribute-based methods present higher accuracy than those based on word vectors~\cite{estevam:2021}.

As representing classes with a set of attributes is not scalable, some recent approaches have replaced attributes by detecting objects in scenes~\cite{jain:2015,mettes:2017}.
This approach works because the visual class-object relationships also exist in texts and are captured in word vectors~\cite{mettes:2017}.
Nevertheless, it has some limitations; for example, it can be difficult to distinguish foreground and background objects or provide a proper representation for these object labels in the semantic space. Additionally, the presence of out-of-context objects produces incorrect~predictions. 

\begin{figure*}
\centering
\includegraphics[width=0.99\linewidth]{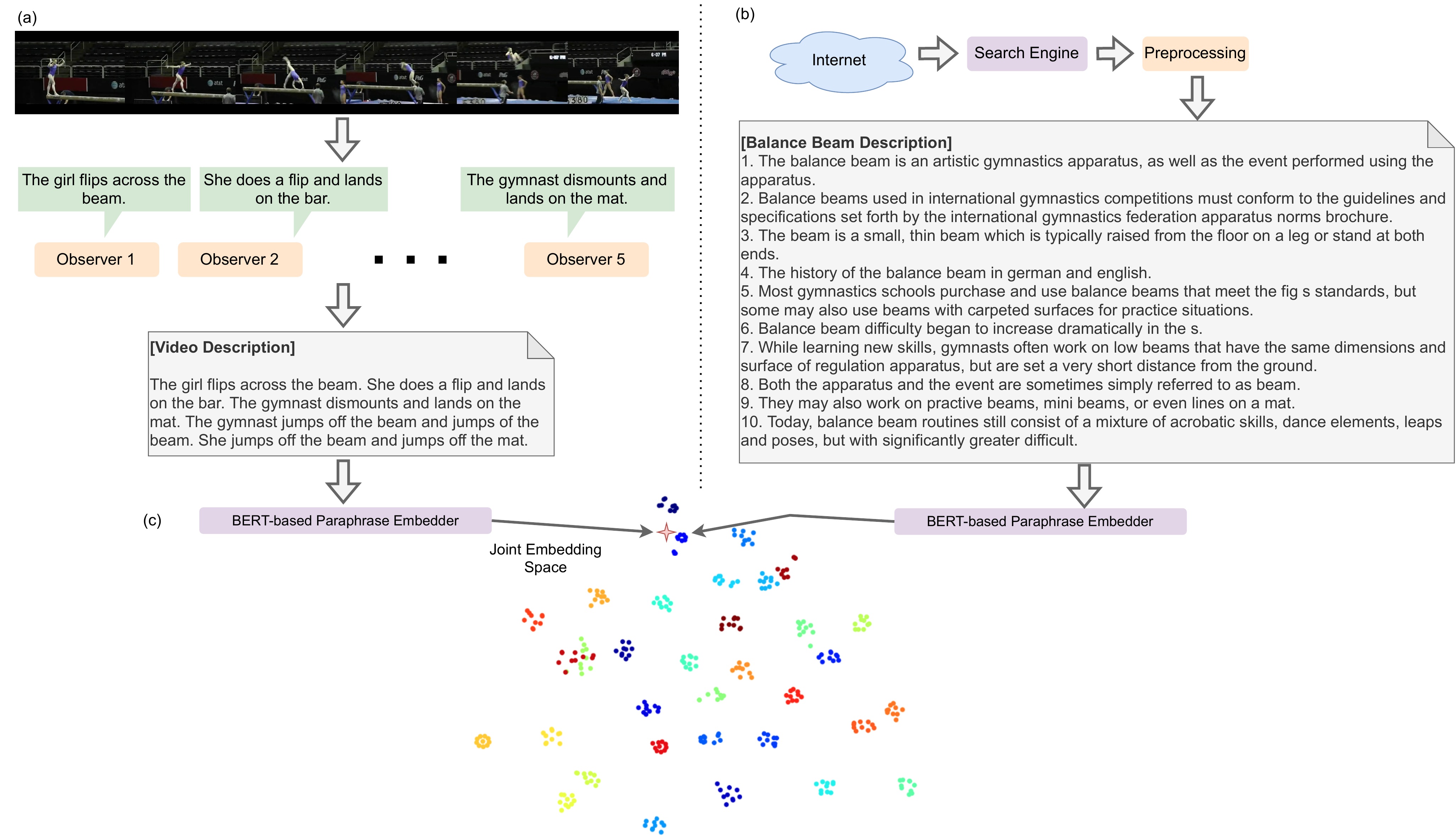}
\caption{\VE{Representation of our ZSAR method. In (a), we show the visual representation procedure. In (b), the semantic representation is shown. Finally, in (c), the joint embedding.}}
\label{fig:method}
\end{figure*}

Considering the above discussion, \major{in this work we propose a method in which the goal is to represent the videos and labels with the same modality of information, aiming to mitigate the domain adaptation problem.
An intuitive choice is to represent labels and videos with sentences or paragraphs in natural language. In that way, we can produce rich representations for both visual and semantic, and our method is illustrated in Fig.~\ref{fig:method}. Although intuitive, this is the first work, to the best of our knowledge, that uses neural networks to convert videos into descriptive sentences, and then to perform \ac{zsar} with these sentences.}

First, we encode the videos using observers that generate a descriptive sentence given an input video, as shown in Fig.~\ref{fig:method}(a). We choose \ac{sota} video captioning architectures from~\cite{iashin:2020,iashin:2020b,estevam:2021b} and pre-training them in the ActivityNet captions dataset (\ie \textit{without any class label}). These architectures present remarkable properties, such as (i)~using self-attention to concentrate on more important segments in the videos; (ii)~storing in their weights video-text relationships; and (iii)~producing fluent sentences, which enable us to estimate the similarity between these sentences and the semantic side information using methods for paraphrase identification.

We then encode the action labels with texts collected from the Internet through search engines, as illustrated in Fig.~\ref{fig:method}(b).
More specifically, we use the descriptions provided by Wang and Chen~\cite{wang:2017b} and employ a simple strategy to select only the sentences most closely related to the action labels. 
We demonstrate this procedure is more effective then those proposed by Chen and Huang~\cite{chen:2021} and our final class description is independent of human evaluation or approval.

As shown in Fig.~\ref{fig:method}(c), we take advantage of \ac{sota} paraphrase methods based on \ac{bert}, and produce a joint embedding space in which a simple \ac{nn} method achieves remarkable~performance.

Our work has some advantages compared to existing methods:
(1)~the semantic gap due to domain adaptation does not exist or is significantly mitigated when comparing a textual video description with a textual class label description;
(2)~a joint latent representation between visual patterns and texts is encoded in video captioning neural networks, being a natural bridge between these information modalities;
(3)~the model is entirely \textit{cross-dataset} and \textit{plug and play}, \ie we can replace the captioning models with others with better performance or trained on other datasets; we can also replace the \ac{bert}-based encoding with an even more accurate encoder with no additional training; and
(4)~ideally, no additional training is required to incorporate more classes.
It is only necessary to collect texts with descriptions for the labels, which can be automated.

Our contributions are summarized as follows:
\begin{enumerate}
    \item We demonstrate that representing videos with descriptive sentences, automatically learned, instead of deep features is viable and conduct us to the \ac{sota} on the UCF101 dataset in the \ac{zsl} scenario;
    \item We demonstrate that class labels encoded with word vectors are unsuitable for building the semantic embedding space for our approach.
    Otherwise, we propose representing the classes with sentences extracted from documents acquired with search engines on the Internet without any human evaluation of their content;
    \item We build a shared semantic space employing a \ac{bert}-based embedder with a highly accurate pre-trained model for the paraphrasing task. 
    The projection onto this space is straightforward for both types of~information;
    \item Finally, our experimental evaluation demonstrated that the main performance limitation is the current state of the art on video captioning, which can be considerably improved in the coming~years by creating new end-to-end models combining these two objectives (captioning and \ac{zsar}).
\end{enumerate}

\section{Related Work}
\label{sec:relatedwork}

The central problem in \ac{zsar} is how to bridge the gap between what the model is seeing and the semantic knowledge it has. As shown in Estevam~\etal~\cite{estevam:2021}, existing methods based on attributes manually annotated reached greater accuracy than raw deep representations.
However, video annotation is not scalable, and different approaches have been proposed to represent videos with automatically detected attributes, usually the presence or absence of objects, classified by knowledge transfer from large-scale datasets.
Recently, the use of textual representations to learn joint representations has been proposed with promising performance.
In the following subsections, we introduce some relevant approaches for these strategies. 

\subsection{Object Representations for ZSAR}

Guadarrama~\etal~\cite{guadarrama:2013} proposed an approach based on hierarchical semantic models for subjects, objects, and verbs.
They employed object detectors associating the predictions with their corresponding leaves in the hierarchies. 
Information from objects and subjects is combined and fed into a non-linear~\ac{svm}. 
On the other hand, Jain~\etal~\cite{jain:2015} used the estimated probability of detected objects as prior knowledge and estimated an affinity between an object class and an acting class.
This information was used to compute the semantic description of an action class as a function of the set of predicted objects. 

Zuxuan~\etal~\cite{zuxuan:2016} proposed generating an intermediate space containing the relationships among objects, scenes, and actions.
They employed a semantic fusion network on three streams: global low-level \ac{cnn} (\eg from a VGG19 trained on ImageNet); object features in frames (\eg from VGG19 trained on a subset of 20{,}574 objects); and features of scenes (\eg from a VGG16 trained on the Places205 dataset).
The correlation between objects/scenes and video classes is mined from the visualization of the network by salience maps producing a matrix with the probability that each pair (object, scene) is related to an action.

Mettes and Snoek~\cite{mettes:2017}, on the other hand, focused on the spatial relationship between actors and objects.
They proposed a method based on spatial-aware object embeddings computed from interactions between actors and local objects in sequential frames using a pre-trained Faster R-CNN model on the MS-COCO dataset.
Segments with actor-local object interaction were called action tubes, and these tubes are distinguished among different videos using global object classifiers through the GoogleLeNet network.
The video class is determined as the class with the highest combined score between video tube embeddings and global classifiers. Their semantic information is given by cosine distance of actions and objects taken Word2Vec~representations.

Gao~\etal~\cite{gao:2019} learned the relationship between actions and objects in a two-stream configuration.
In the first stream, they learned classifiers on graph models constructed with ConceptNet5.5~\cite{speer:2017}, where the concepts are represented with word vectors. 
The second stream used the visual representations of objects (with the methods used in~\cite{jain:2015} and~\cite{mettes:2017}) to learn the graphs.
The classifiers are learned during training and optimized for seen categories.
Hence, in testing, the classifiers of unseen categories (\ie from the first stream) are used to classify the object features of test videos (\ie from the second stream).

\VE{Ghosh~\etal~\cite{ghosh:2020} were inspired by~\cite{gao:2019}. In their work, knowledge graphs were fed} to a~\ac{gcn}, aiming to minimize the \ac{mse} between the final classifier layer weights (\ac{gcn}) with the classifier layer weights from~\ac{i3d}.

Finally, Kim~\etal~\cite{kim:2021} proposed generating semantic embedding spaces based on dynamic attributes signatures.
They showed that dynamic attributes are preferable to static ones for modeling actions due to the lack of temporal information.
Thus, they constructed finite state machines over the static annotations provided in the UCF101 and Olympic Sports datasets describing the presence and the transitions between these states. 
These patterns are action signatures used to perform the \ac{zsar}~classification.

Our method explores the ability of video captioning to identify objects in scenes inferred by their context and by sentence annotations. Additionally, we employ the \ac{i3d} model as a deep representation, and this model incorporates the weights of an Inception-V1 model pre-trained on ImageNet~\cite{carreira:2017}.

\subsection{Text Representations for ZSAR}

Zhang~\etal~\cite{zhang:2018} proposed an improved model for learning visual and textual alignments.
Typically, these approaches take a set of paragraphs, represented as a sequence of words, and feed it into an encoder to obtain a paragraph embedding.
Similarly, a set of short clips composed of a few frames is fed to an encoder to obtain a video embedding.
These embeddings are updated with a loss function at a high level (\eg cosine distance). Their method proposes a mid-level alignment where paragraphs are aligned to videos and sentences are aligned to short clips.
The quality of the intermediate encoding is improved by using decoding networks to evaluate reconstruction errors.

Piergiovanni and Ryoo~\cite{piergiovanni:2020} also developed a method to learn an intermediate representation for both videos and texts based on an encoder-decoder approach.
In their method, there are two encoder-decoder pairs: (video-encoder, video-decoder) and (text-encoder, text-decoder).
The first encoder takes a video and produces an intermediate space, and the first decoder reconstructs the video given the intermediate representation.
The same occurs with text.
Four loss functions were proposed to handle the learning with paired and unpaired data.
The classification is performed by the \ac{nn} rule between each video representation and its text representation in the intermediate~space.

Recently, Chen and Huang~\cite{chen:2021} proposed a method combining object detection and textual information.
They observed that only word vector representation is insufficient to provide information for objects detected in the videos. 
Then, they used the object label to retrieve their WordNet description as an object concept description.
Additionally, they proposed a combination of Wikipedia and dictionary data to compose action class descriptions using human supervision in this task. Hence, they could identify objects in videos and provide a representation based on their concepts.
Although well succeeded, their method requires the presence of visual representation in the \ac{zsar} classification~step. 

Our method is also based on textual descriptions, but it has several differences:
(1)~we use methods that predict descriptions word by word and consider the visual information and the previously predicted words.
A clear advantage of this strategy is to ignore objects out of context;
(2)~our method does not require any class label annotation nor to train the~\ac{zsar} classifier;
(3)~our strategy for semantic side representation does not require human supervision at the level of sentences; 
it requires only a document from the Internet with a general description; and
(4)~as we have good descriptions, paraphrase identification methods pre-trained on millions, or even billions of sentences, can be employed without the need for~fine-tuning.

\section{Methodology}
\label{sec:methodology}

In this section, we describe in detail our methodology, which is illustrated in Fig.~\ref{fig:method}.
\VE{To facilitate our presentation, Table~\ref{tab:tableofsymbols} summarizes the notations used in this paper.}

\input{tables/tableofsymbols}

\subsection{Problem Definition}

The goal of \ac{zsar} is to classify samples belonging to a set of unseen action categories $ \mathcal{Y}_{u} = {y_{1},...,y_{u_{n}}} $ (\ie never seen before by the model) given a set of seen categories $ \mathcal{Y}_{s} = {y_{1},...,y_{s_{n}}} $ as the training set. The problem is named ZSAR only if the following restriction is respected:
\begin{equation}
    \label{eq:zslassumption}
    \mathcal{Y}_{u} \cap \mathcal{Y}_{s} = \emptyset
\end{equation}

Our classification consists of mapping both video and semantic information (\ie class description) into a joint embedding space. Then, the classification is performed with a \ac{nn} rule under some similarity function, such as
\begin{equation}
    \label{eq:zslclassification}
    y_{pred} = \argmax_{y_{prot} \in \mathcal{Y}_{u_{prots}}} \textit{Sim}(\textit{Emb}(y_{prot}),\textit{Emb}(Obs(v)))
\end{equation}
\noindent \VE{in which $\textit{Sim}(\cdot)$} is the cosine similarity; $v$ is a video, $Obs(\cdot) = [Ob_{1}(\cdot),...Ob_{o}(\cdot)]$; $[\cdot]$ is a concatenation operator and $Ob(\cdot)$ is a video sentence description from each of the $o$ observers (\ie video captioning methods) (see details in Section~\ref{subsec:videorepresentation}); $y_{prot}$ is a sentence from a large textual description for each class obtained with the  procedure described in Section~\ref{subsec:classlabelrepresentation}; finally, $\textit{Emb}(\cdot)$ is a sentence embedding function described in Section~\ref{subsec:sentencesembedding}.
Our method, as mentioned previously, does not use the training set because the benchmark datasets do not provide annotated sentences for their~videos.

\subsection{Video Representation}
\label{subsec:videorepresentation}

\begin{figure*}
\centering
\includegraphics[width=0.90\linewidth]{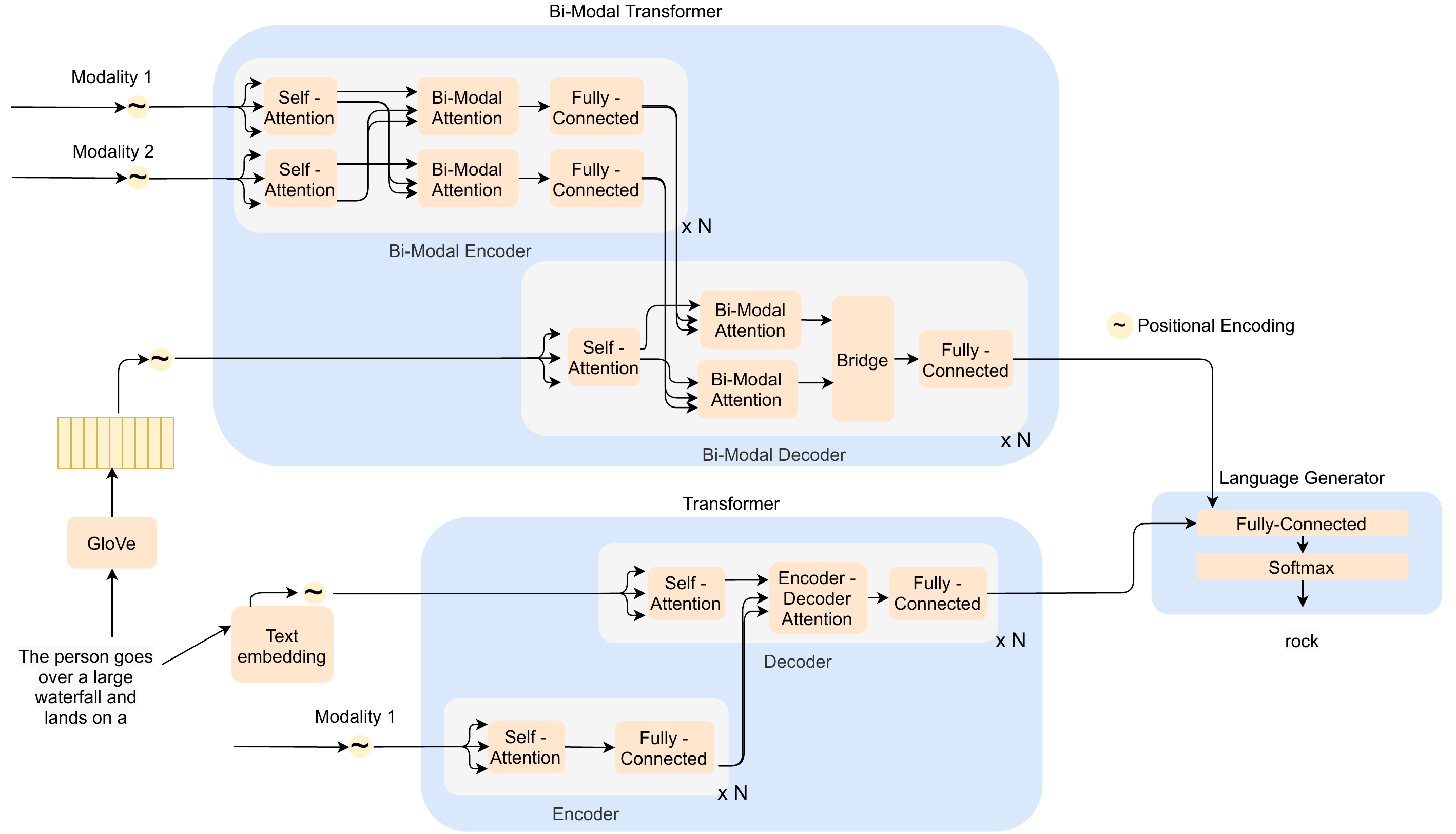}
\caption{\VE{Overview of the captioning architectures showing the BMT and Transformer layers with their inputs and the language generation module. Adapted from~\cite{estevam:2021}.}}
\label{fig:architecture}
\end{figure*}

Our goal is to predict a sentence given a video (using visual and audio information when available).
As video captioning is an area of computer vision responsible for study models with this ability, we choose two \ac{sota} architectures that could be used with the same set of features: Transformer~\cite{iashin:2020} (using the original transformer implementation from \cite{vaswani:2017}), and Bi-Modal Transformer (BMT)~\cite{iashin:2020b}. Fig.~\ref{fig:architecture} shows a diagram illustrating both models.

\noindent \textbf{Transformer:}
First, given a video $v$, the observer takes a set of $n_c$ visual features $v_{f}=\{v_{f_1},...,v_{f_{n_c}}\}$, one per each frame stack, and a set of $m$ words $Y=\{y_1,...,y_m\}$ to estimate the conditional probability of an output sequence given an input~sequence. 

We encode $v_{f_c}$, where $1 \leq c \leq n_c$ as 
\begin{equation}
\label{eq:visualfeature}
v_{f_c}= V_{E}(v_{c}) \, ,
\end{equation}
\noindent where $V_{E}(\cdot)$ yields a deep representation given by an off-the-shelf convolutional network, and $v_c$ is the $c$-th frame stack for the video~$v$.
The video features (Equation~\ref{eq:visualfeature}) are fed all at once to the transformer encoder in which a learned continuous representation is passed to a decoder to generate a sequence of symbols~$Y$ from the language vocabulary.

The Transformer requires information on the position of each feature, and a usual strategy is to compute a positional encoding with sine and cosine at different frequencies as
\begin{eqnarray}
    \label{eq:poswise}
    \textit{PE}_{(pos, 2i)} & = & \sin{(pos/10000^{2i/d_{model}})}, \\
    \textit{PE}_{(pos, 2i+1)} & = & \cos{(pos/10000^{2i/d_{model}})}, \nonumber
\end{eqnarray}
\noindent where $pos$ is the position of the visual feature in the input sequence, $0 \leq i < d_{model}$ and $d_{model}$ is a parameter defining the internal embedding dimension in the transformer.
Subsequently, a multi-head attention layer processes these representations with scaled dot-product attention defined in terms of queries ($Q$), keys ($K$), and values ($V$) as
\begin{equation}
    \label{eq:attention}
    \textit{Att}(Q,K,V)=\textit{softmax}(\frac{{Q} \times K^{T}}{\sqrt{d_{k}}}) \times V,
\end{equation}
\noindent and the multi-head attention layer is the concatenation of several heads ($1$ to $h$) of attention applied to the input projections (computed with dense layers) as
\begin{equation}
    \label{eq:multihead}
    \textit{MHAtt}(Q,K,V)=[head_{1},..., head_{h}]\times W^{0} \, ,
\end{equation}
\noindent where $\textit{head}_{i}=\textit{Att}(Q \times W_{i}^{Q},K \times W_{i}^{K},V \times W_{i}^{V})$ and $[\text{ }]$ is a concatenation operator.
The key insight on Transformer is the self-attention, which takes $Q = K = V = V_{f}^{\textit{PE}}$, resulting in 
\begin{equation}
\begin{split}
V_{f}^{self-att}=[\textit{Att}(V_{f}^{\textit{PE}} \times W_{i}^{V_{f}^{\textit{PE}}},V_{f}^{\textit{PE}} \times W_{i}^{V_{f}^{\textit{PE}}},V_{f}^{\textit{PE}} \times W_{i}^{V_{f}^{\textit{PE}}}), \\
...,\textit{Att}(V_{f}^{\textit{PE}} \times W_{h}^{V_{f}^{\textit{PE}}},V_{f}^{\textit{PE}} \times W_{h}^{V_{f}^{\textit{PE}}},V_{f}^{\textit{PE}} \times W_{h}^{V_{f}^{\textit{PE}}})].    
\end{split}
\end{equation}

The latent feature from the encoder is given by a fully connected feed-forward network $\textit{FFN}(\cdot)$ applied to each position separately and identically, defined as
\begin{equation}
    \label{eq:ffn}
    \begin{split}
    \textit{FFN}(u) = \max(0, u \times W_{1}+b_{1}) \times W_{2}+b_{2} 
    \end{split} \; ,
\end{equation}
\noindent resulting in $V_{f}^{\textit{FFN}}$, which is a rich video representation based on self-attention used in the decoder layer.

The decoder layer receives words and feeds an embedding layer $\textit{E}(\cdot)$, computing the position with Equation~\ref{eq:poswise} resulting in $W^{\textit{PE}}$. This representation is fed to the multi-head self-attention layer to compute an internal representation based on self-attention applied on word sequence, resulting in $W^{self-att}$.

Then, we compute the relationship between video and sentence by feeding the encoder-decoder attention layer, resulting in an attention on the words given the visual encoding as
\begin{equation}
\label{eq:multiheaddecoder}
W^{VisAtt} = \textit{MHAtt}(W^{self-att}, V_{f}^{\textit{FFN}},V_{f}^{\textit{FFN}}).
\end{equation}

Finally, $W^{VisAtt}$ feeds an $\textit{FFN}(\cdot)$ and, then, a generator $G(\cdot)$ composed of a fully connected layer and a softmax layer is responsible for learning the predictions over the vocabulary distribution~probability. This model is highly efficient in modeling visual-textual~relationships.

\noindent \textbf{\ac{bmt}:}
The second architecture employed is \ac{bmt}. Considering the encoder, this transformer has two differences from the Transformer encoder.
It takes two streams, visual $V_{f}$ and audio $\textit{A}$ \cite{iashin:2020b} or semantic $\textit{Sm}$ \cite{estevam:2021b}, separately. We denote this second stream as $\textit{ASm}$ (\ie audio or semantic). The encoder has three sub-layers: self-attention (Equation~\ref{eq:attention}), producing $V_{f}^{self-att}$ and $\textit{ASm}^{self-att}$; 
bi-modal attention, i.e.,
\begin{eqnarray}
V_{f}^{\textit{ASm}-att} & = & \textit{MHAtt}(V_{f}^{self},\textit{ASm}^{self}, \textit{ASm}^{self}),
\label{eq:visualsemanticattended} \\
\textit{ASm}^{Vis-att} & = & \textit{MHAtt}(\textit{ASm}^{self}, V_{f}^{self},V_{f}^{self}),
\label{eq:semanticvisualattended}
\end{eqnarray}
\noindent and a fully connected layer $\textit{FFN}(\cdot)$ for each modality attention, producing $V_{{\textit{ASm}-att}}^{\textit{FNN}}$ and $\textit{ASm}_{v-att}^{\textit{FNN}}$ used in the bi-modal attention units on the decoder.

Considering the bi-modal decoder, a $W^{self-att}$ is obtained with Equation~\ref{eq:multihead}.
Afterward, the bi-modal attention is computed as
\begin{equation}
\label{eq:wordsmatt}
W^{\textit{ASm}-att}=\textit{MHAtt}(W^{self-att}, \textit{ASm}_{v-att}^{\textit{FNN}},\textit{ASm}_{v-att}^{\textit{FNN}}) \, ,
\end{equation}
\noindent and
\begin{equation}
\label{eq:wordvisatt}
W^{V-att}=\textit{MHAtt}(W^{self-att}, V_{{\textit{ASm}-att}}^{\textit{FNN}},V_{{\textit{ASm}-att}}^{\textit{FNN}}).
\end{equation}

The bridge is a fully connected layer on the concatenated output of bi-modal attentions, which are enriched features through attention on the combination of two video modalities (\eg visual and audio), computed as
\begin{equation}
\label{eq:z}
W^{\text{FFN}}=\text{FFN}([W^{Sm-att}, W^{V-att}]).
\end{equation}

The output of the bridge is passed through another $\textit{FFN}$ and then to the generator $G(\cdot)$.
This means that the encoder parameters are learned conditioning them to the sentence output quality. 

We compute the semantic descriptor from~\cite{estevam:2021b} strictly following the model and training procedures. The mathematical details can be found in the original~paper.

\subsection{Class Label Representation}
\label{subsec:classlabelrepresentation}

We take a dataset with documents collected on the Internet containing a textual description for each class. Hence, for each class, we have a set of prototype sentences $\mathcal{Y}_{\text{prot}} = \{y_{\text{prot}_{1}}, y_{\text{prot}_{2}}, ..., y_{\text{prot}_{q}}\}$ obtained by splitting the paragraphs. 

We employ simple but effective selection criteria: (i) to filter the sentences with a minimum number of words; (ii) to compute dense representations for all the sentences and the class label using the \ac{sbert}~\cite{reimers:2019} model; (iii) to compute the cosine similarity between the dense representations of the class label and the sentences; and (iv) to select a maximum number of sentences ordered by the highest similarity.

The joint embedding space used for~\ac{zsar} is composed of representations for video and prototype sentences computed with the SBERT model. The details are provided in the following section.

\subsection{Sentence Embedding}
\label{subsec:sentencesembedding}

\begin{figure}[!htb]
    \centering
    \captionsetup[subfigure]{captionskip=-0.25pt,font={scriptsize},justification=centering} 
    \subfloat[][]{
    \resizebox{0.48\linewidth}{!}{
    \includegraphics[height=7ex]{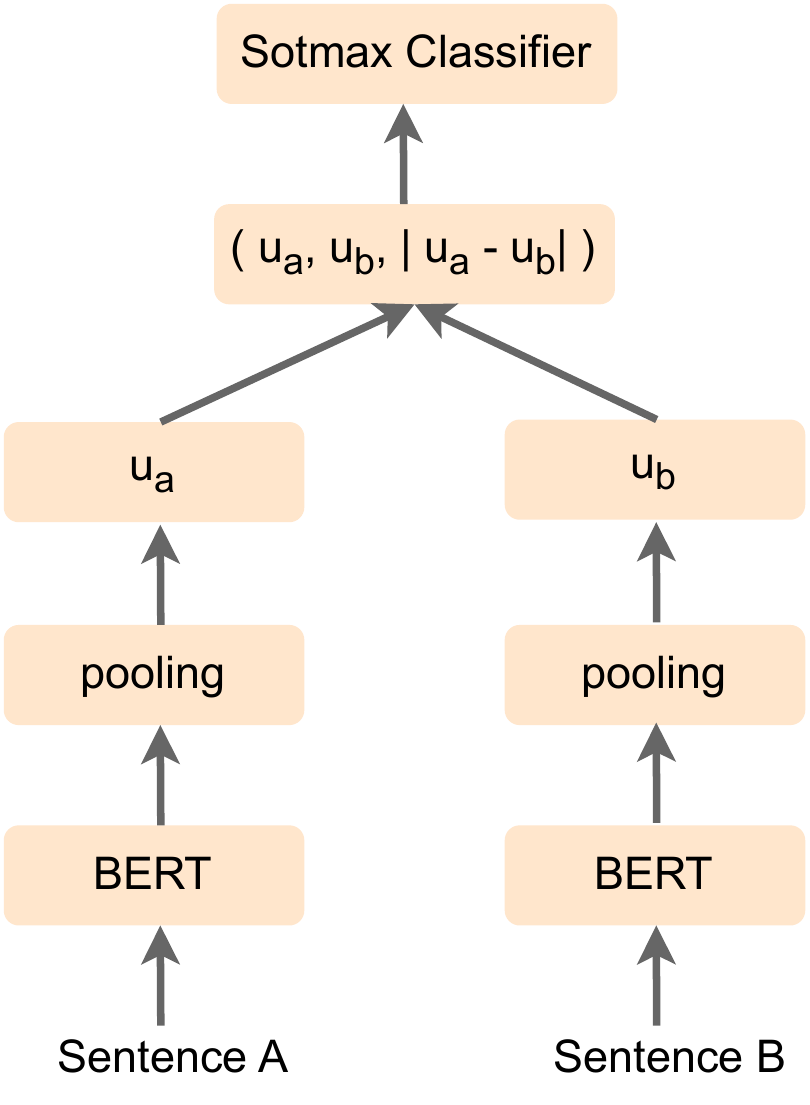}
    }}
    \subfloat[][]{
    \resizebox{0.48\linewidth}{!}{
    \includegraphics[height=7ex]{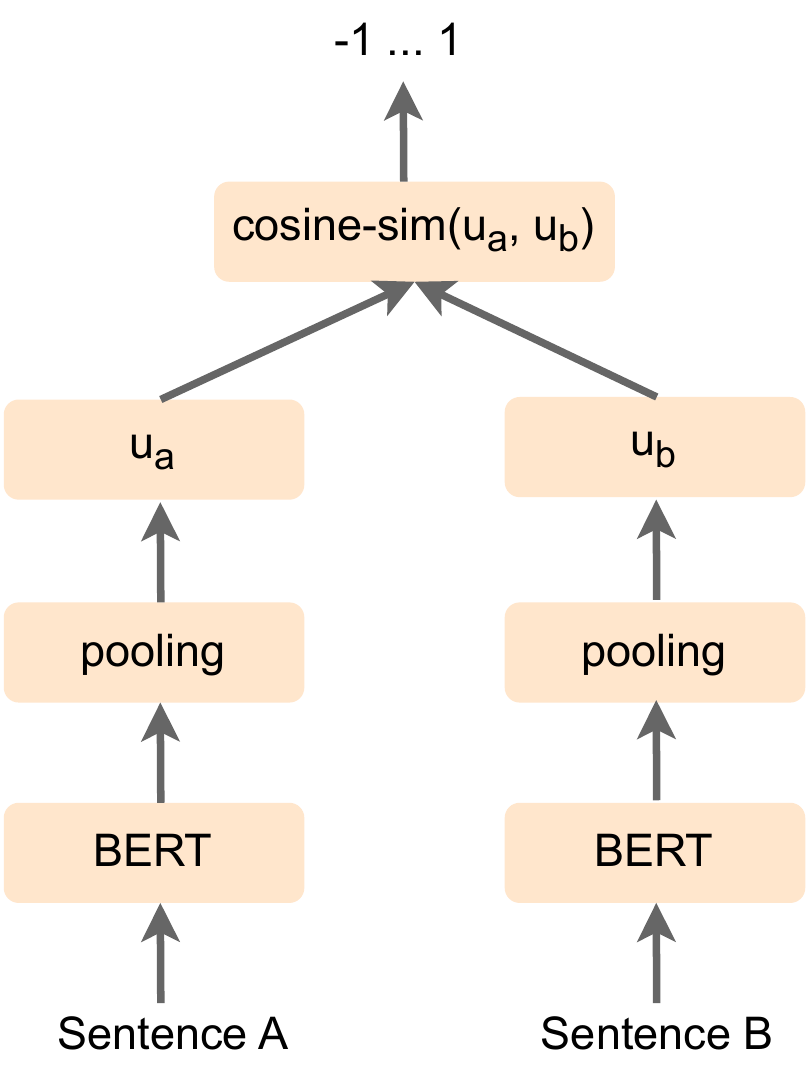}
    }
    }
    \vspace{-0.25mm}
    \caption{SBERT architecture from Reimers and Gurevych~\cite{reimers:2019}. In (a) is shown the classification objective function, and in (b) the architecture used at the inference or regression tasks.}
    \label{fig:sbertarchitecture}
\end{figure}

We propose to encode information at the level of sentences and not words. For this task, we use the SBERT model from~\cite{reimers:2019}. It is an improved \ac{bert}~\cite{devlin:2019} model that drastically reduces the computational cost for acquiring \ac{bert} embeddings by feeding a Siamese network, containing two~\ac{bert} models, with one sentence per branch, dispensing with the special token [SEP].
The model architecture is shown in Fig.~\ref{fig:sbertarchitecture}.

\ac{bert} or RoBERTa models are fine-tuned on large-scale textual similarity datasets. If the dataset requires classification, the objective function is described as
\begin{equation}
    \label{eq:sbertclassificationobjective}
    o = \textit{softmax}(W_{t}(u_a, u_b, \lvert u_a - u_b \rvert)))
\end{equation}
\noindent where $\lvert u_a - u_b \rvert$ is an element-wise subtraction, $W_{t}\in \mathbb{R}^{3n_s \times k}$ is the trainable weights, $n_s$ is the dimension of sentence embeddings, and $k$ is the number of labels. The model optimizes the cross-entropy loss. On the other hand, if the dataset requires regression, the cosine similarity between two sentence embeddings $u_a$ and $u_b$ is computed, and the loss function is the mean squared~error.

The model can also be optimized using a triplet objective function. Taking an anchor sentence $a$, a positive sentence $p$, and a negative sentence $n$, the triplet loss tunes the network so that the distance between $a$ and $p$ is smaller than the distance between $a$ and $n$, that is, minimizing the following equation
\begin{equation}
    \label{eq:tripletloss}
    max(\parallel s_a - s_p \parallel - \parallel s_a - s_n \parallel + \epsilon, 0),
\end{equation}
\noindent where $s_a$, $s_p$, and $s_n$ are sentence embeddings, $||\cdot||$ is a distance metric and $\epsilon$ is a margin ensuring that $s_p$ is at least $\epsilon$ closer to $s_a$ than $s_n$.

Our interest is in the vector $u$ (see Fig.~\ref{fig:sbertarchitecture}), after the fine-tuning, computed as the mean of all outputs instead only output for [CLS], as occurs in BERT. For details on BERT or RoBERTa see~\cite{devlin:2019} and \cite{liu:2019}, respectively.

\section{Experiments}
\label{sec:experiments}

In this section, we introduce the datasets and protocols, the implementation details, and the results. We also include an extensive ablation study organized as a set of questions and answers (\textit{Q\&A}).

\subsection{Datasets and \VE{Protocols}}

Our observers were trained using the ActivityNet Captions dataset~\cite{krishna:2017}, which consists of $10{,}024$ training, $4{,}926$ validation, and $5{,}044$ testing videos collected from YouTube. The videos are annotated with start and end points for events, and a sentence is provided for each annotation totaling approximately $36$K pairs of event-sentence. The sentences have an average length of $16.5$ words and describe around $36$s of their videos. It is important to highlight that no action label from ActivityNet is used during the training of the video~observers.

\begin{figure*}[!htb]
    \centering
    \resizebox{0.99\linewidth}{!}{
    \includegraphics{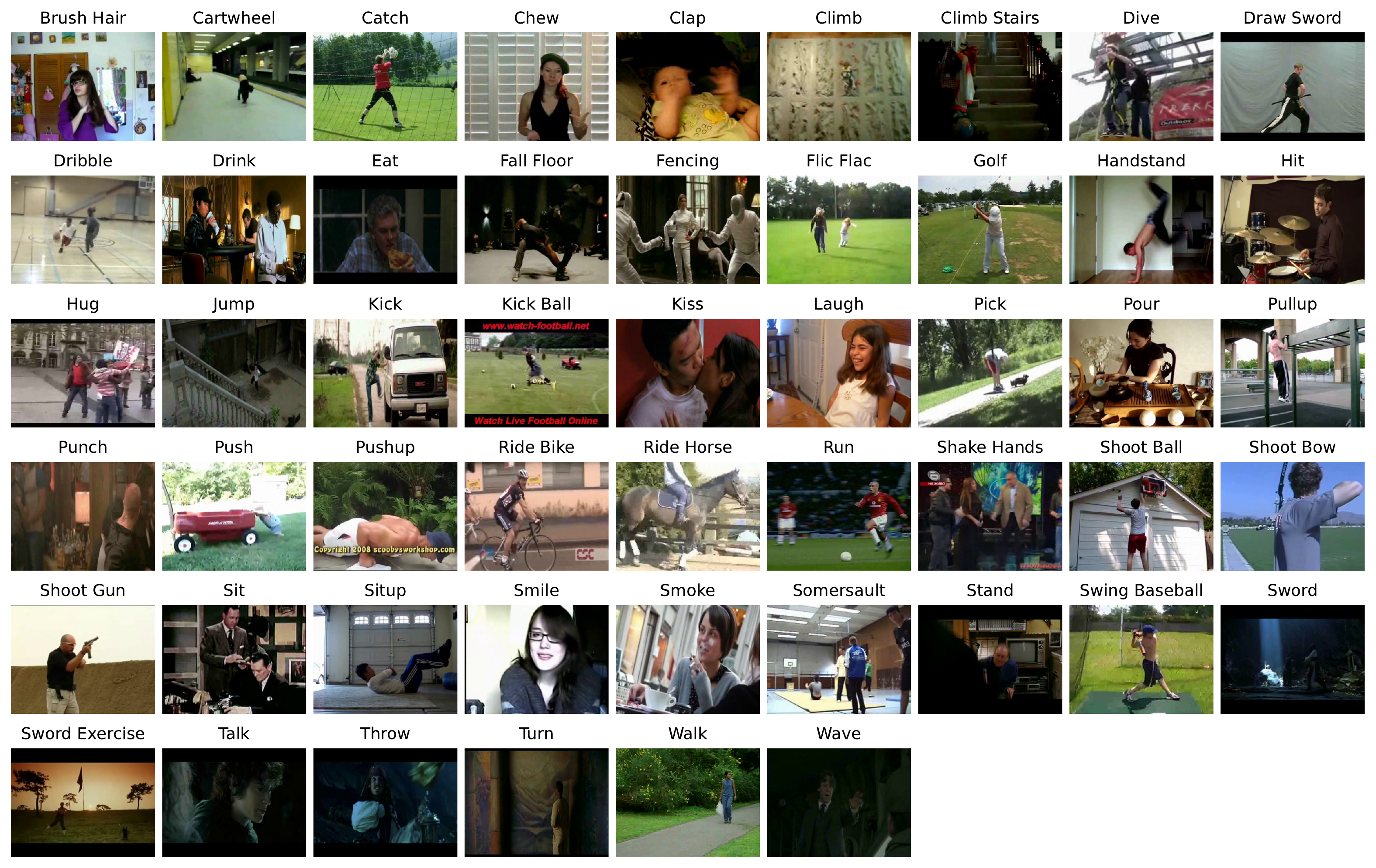}
    }    
    \caption{\VE{Samples for the 51 action classes from the HMDB51 dataset~\cite{kuehne:2011}.}}
    \label{fig:samplesfromhmdb51}
\end{figure*}

\begin{figure*}[!htb]
    \centering
    \captionsetup[subfigure]{captionskip=-0.25pt,font={scriptsize},justification=centering} 
    \resizebox{0.99\linewidth}{!}{
    \includegraphics{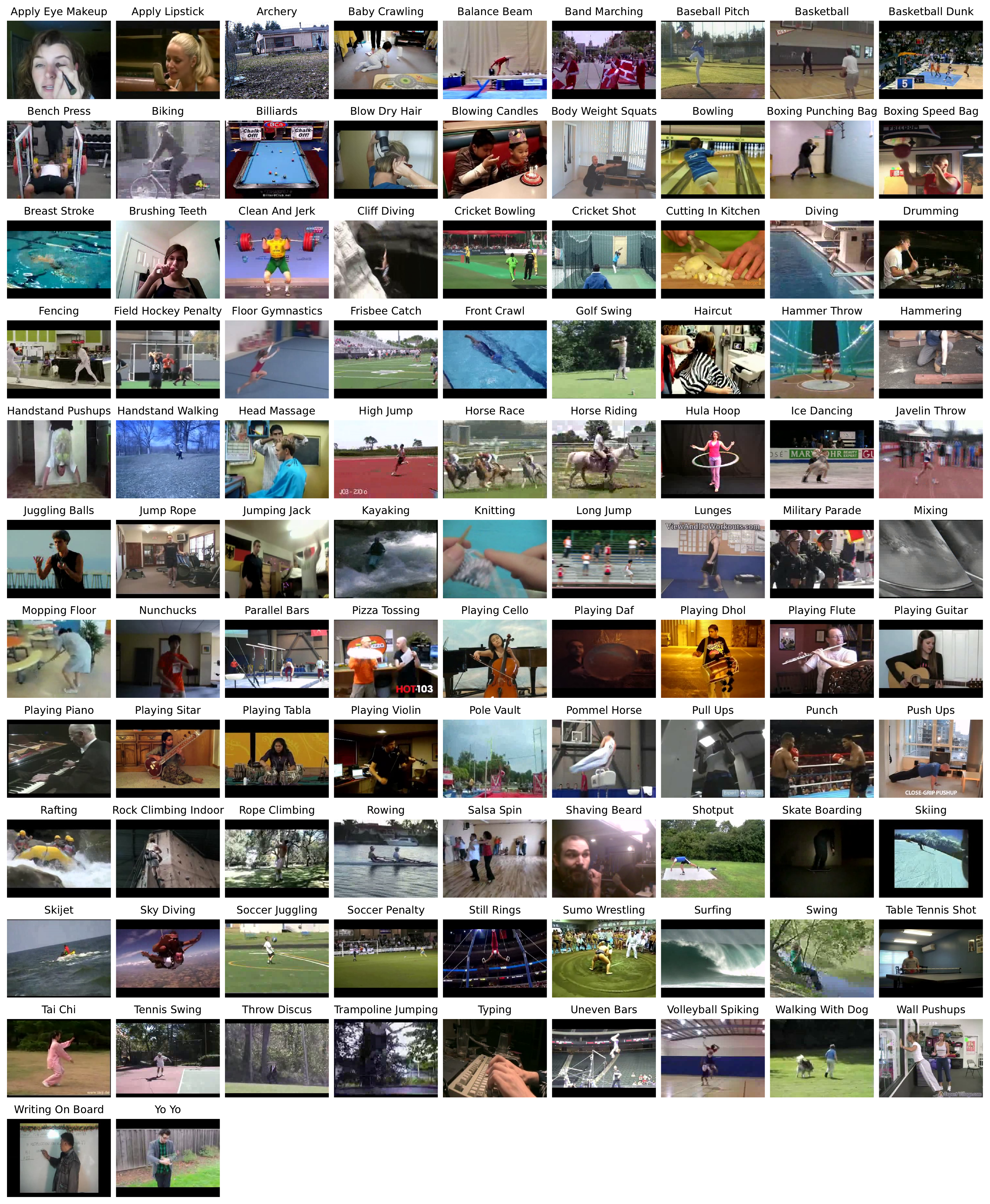}
    }        
    \vspace{0.5mm}
    \caption{\VE{Samples for the 101 action classes from the UCF101 dataset~\cite{soomro:2012}.}}
    \label{fig:samplesfromucf101}
\end{figure*}

For testing, we employ the popular benchmarks HMDB51~\cite{kuehne:2011} and UCF101~\cite{soomro:2012}. The former is composed of $6{,}766$ videos from $51$ classes, \VE{illustrated in Fig.~\ref{fig:samplesfromhmdb51}}, with an average duration of $3.2$s; the frame height is scaled to $240$, and the frame rate is converted to $30$ \ac{fps}. The latter comprises $13{,}320$ videos from $101$ action classes, \VE{illustrated in Fig.~\ref{fig:samplesfromucf101}}, with frame resolution standardized to $25$ \ac{fps} and $320\times240$ pixels.
The average duration of the videos is $7.2$s. \major{As is customary in \ac{zsar} research~\cite{estevam:2021}, performance is evaluated using the well-known accuracy metric, which quantifies the number of correct predictions relative to the total number of predictions~made}.

Providing a fair evaluation of \ac{zsar} models using these datasets is not straightforward due to the nature of the visual feature extractors and the datasets used for training them. For example, \major{if a \ac{zsar} model uses the \ac{i3d} network, pre-trained on Kinetics400~\cite{carreira:2017},  there are overlaps between the set of classes from Kinetics400 and the set of classes from HMDB51 and UCF101. This overlap imposes the removal of these classes from the \ac{zsar} test set to preserve the \ac{zsl} premise (\ie the disjunction between training and testing class sets).} However, these overlaps are often challenging to recognize due to differences in class names and the visual and semantic similarity between certain classes, as pointed out in~\cite{estevam:2021,brattoli:2020,roitberg:2018,chen:2021,gowda:2021b}. 

\major{Taking this into account}, we adopt the TruZe evaluation protocol~\cite{gowda:2021b} on UCF101 and HMDB51 datasets, in which the testing split is generated with the following guidelines: (i)~to discard exact matches (\eg archery); (ii)~to discard matches that can be either superset or subset (\eg cricket shot and cricket bowling (UCF101) and playing cricket (Kinetics400)); and
(iii)~to discard matches that predict the same visual and semantic match (\eg apply eye makeup (UCF101) and filling eyebrows (Kinetics400)).
The result is a configuration with $29/22$ (train/test) and $67/34$ classes for the HMDB51 and UCF101 datasets, respectively.
As our model does not require these training sets (\ie it is cross-dataset), we take into consideration only the testing sets (\ie $0/22$ and $0/34$)\footnote{\textbf{UCF101} - apply lipstick, balance beam, baseball pitch, billiards, blow dry hair, cutting in kitchen, fencing, field hockey penalty, front crawl, hammering, handstand pushups, handstand walking, horse race, ice dancing, jumping jack, military parade, mixing, nunchucks, parallel bars, pizza tossing, playing daf, playing dhol, playing sitar, playing tabla, pommel horse, punch, rafting, rowing, still rings, sumo wrestling, table tennis shot, uneven bars, wall pushups, and yo yo; \textbf{HMDB51} - chew, climb stairs, draw sword, fall floor, fencing, flic flac, handstand, hit, jump, kick, pick, pour, run, sit, shoot gun, smile, stand, sword exercise, talk, turn, walk, and~wave.}.

\VE{Finally, we also provide a comparison using a conventional protocol employed in most of the works. 
In some cases 0/50\%, and in the most 50\%/50\%\footnote{Not all methods allow 0/50 experiments.}. Although there are overlaps in training and testing sets, several methods employ this scheme~\cite{wang:2017b,mandal:2019,lee:2021,sun:2022}. This evaluation is important to observe the impact of the use of i3D features on the results and how our method compares to others independently of the adopted protocol.}

\subsection{Implementation Details}

\begin{figure*}[!htb]
    \centering
    \captionsetup[subfigure]{captionskip=-0.25pt,font={scriptsize},justification=centering} 
    \subfloat[][]{
    \resizebox{0.61\linewidth}{!}{
    \includegraphics[height=13ex]{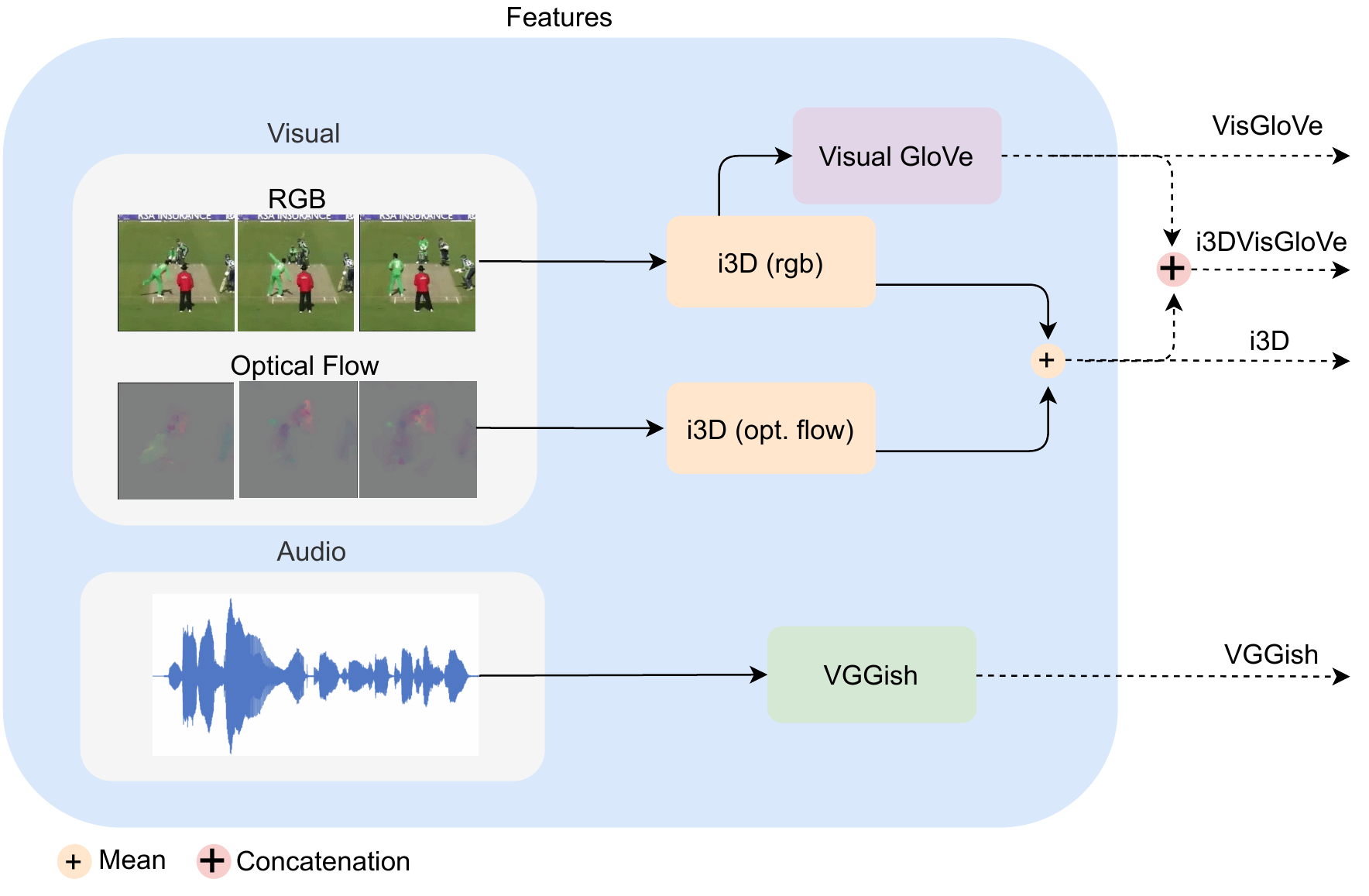}
    }}
    \subfloat[][]{
    \resizebox{0.35\linewidth}{!}{
    \includegraphics[height=3.5ex]{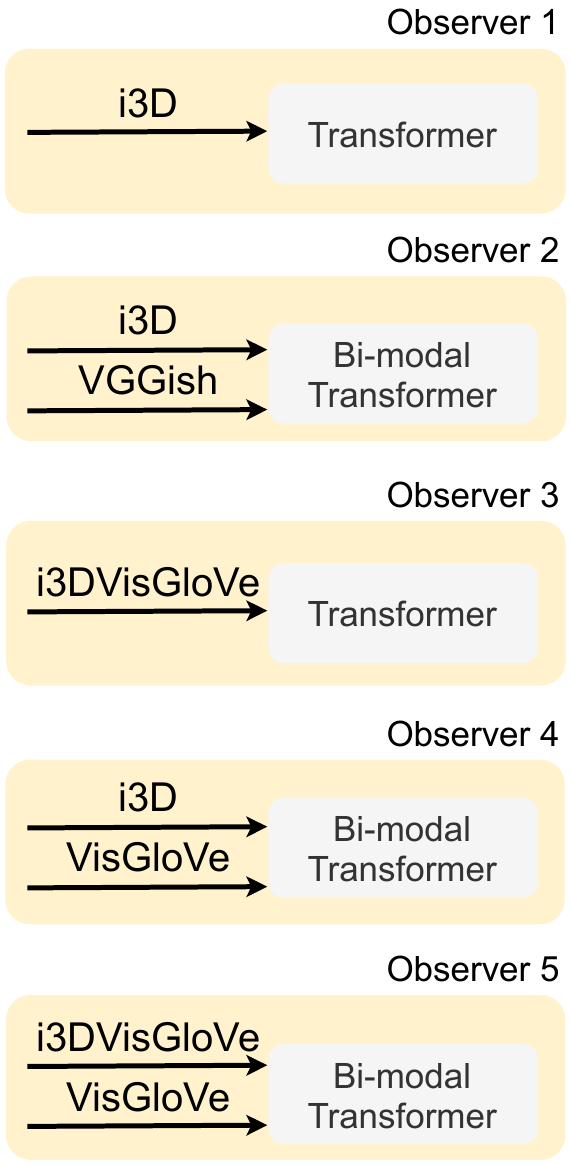}
    }
    }
    \vspace{-0.25mm}
    \caption{Features and observers. In (a) is shown features computed from visual and audio streams, and in (b) the observers architecture and their respective input features.}
    \label{fig:featuresandobservers}
\end{figure*}

We compute features as shown in Fig.~\ref{fig:featuresandobservers}.
For all videos, we extract features from all datasets using the \ac{i3d} network with its two streams, RGB and Optical Flow, in videos with $25$ \ac{fps}. We follow the authors' recommendations for re-scaling ($224\times224$ pixels) but replace the TV-L1~\cite{mohamed:2012} optical flow algorithm for the PWC-Net~\cite{sun:2017}, as it is much faster\footnote{The code used for feature extraction is available at \url{https://github.com/v-iashin/video_features}}.
For each video, we extract one feature with stacks of $24$ frames and steps of $24$ frames (\ie $0.96$ features per second). The audio features are extracted with the VGGish model~\cite{hershey:2017} pre-trained on AudioSet~\cite{gemmeke:2017}.
We follow the default configuration.

Considering that the videos on the HMDB51 dataset do not have the audio signal and that around $50$\% of the videos from UCF101 have this information, we compute the Visual GloVe features~\cite{estevam:2021b} from RGB stream of \ac{i3d}, which is a simple and effective feature to replace the audio stream in the \ac{bmt} model and to enrich the Transformer model input. Finally, we get four features: VisGloVe, i3DVisGloVe, i3D, and VGGish (see Fig.~\ref{fig:featuresandobservers}(a)). 
With these features, we fed two architectures for video captioning (\ie Transformer and \ac{bmt}) which allowed us to generate 5 distinct observers. Fig.~\ref{fig:featuresandobservers}(b) shows the configuration of each observer (architecture and inputs). 

The Transformer and \ac{bmt} models are trained up to $60$ epochs employing early stopping if the Meteor score~\cite{banerjee:2005} stays unchanged for $10$ epochs. The loss function adopted is the Kullback-Leibler Divergence with label smoothing and masking. Dropout is used to prevent overfitting with a rate of $0.1$. Additionally, we monitor the Bleu@3 and Bleu@4 scores~\cite{papineni:2002} to allow evaluating the quality of the sentences produced during the training stage.
The \ac{visglove} features are computed with a vocabulary of $1{,}000$ visual words (learned with clustering), a context of $25$ words ($\approx$~$24$s), and a dimension of $128$. The training is performed until $1{,}500$ epochs with early stopping of $100$ without improvements in the cost function.

The adoption of multiple observers is motivated by the intuition that different humans would produce different sentences given a sample video. Although different, these sentences would tend to be complementary to each other. As our results show, this scheme is highly efficient in improving the video representation, which is reflected in the increase of \ac{zsar} accuracy considering multiple~sentences.

\VE{We use the textual descriptions provided in~\cite{wang:2017b}\footnote{The data is available at \url{https://staff.cs.manchester.ac.uk/~kechen/ASRHAR/}} as side information. The texts are processed using the NLTK\footnote{\url{https://www.nltk.org/}} package for splitting paragraphs into sentences and the \textit{contractions}\footnote{\url{https://pypi.org/project/contractions/}} package to expand contractions (\eg ``isn't'' to ``is not''). We follow the procedure described in Section~\ref{subsec:classlabelrepresentation} by selecting sentences with a minimum of $10$ words and up to $10$ sentences per class and taking the nearest sentence encodings (cosine similarity) compared to the label encoding. The sentences from the observers are concatenated. We build the joint space with Sentence-BERT encoders~\cite{reimers:2019}, namely, the \textit{paraphrase-distilroberta-base-v2}\footnote{Trained on the following datasets: AllNLI, sentence-compression, SimpleWiki, altlex, msmarco-triplets, quora\_duplicates, coco\_captions, flickr30k\_captions, yahoo\_answers\_title\_question, S2ORC\_citation\_pairs, stackexchange\_duplicate\_questions, wiki-atomic-edits.} model~\cite{reimers:2020}. A \ac{nn} algorithm employing cosine distance is used to conduct the ZSAR classification}.

\major{The deep learning models were implemented using PyTorch\footnote{\url{https://pytorch.org/}}, while the ZSAR classifier was implemented using scikit-learn\footnote{\url{https://scikit-learn.org/}}.
All experiments were conducted on a computer system equipped with an AMD Ryzen 7 2700X 3.7GHz CPU, 64 GB of RAM, and an NVIDIA Titan Xp GPU (12 GB).
The experiments were executed on the Ubuntu operating system.}

\subsection{Selected Benchmarks and Evaluation}

We selected two generic~\ac{zsl} models and five~\ac{sota}~\ac{zsar} methods for TruZE comparison, briefly described in this section.

Latem~\cite{xian:2016} is a direct projection onto semantic space method in which a piece-wise linear compatibility function is used to understand the visual-semantic embedding relationships.
SYNC~\cite{changpinyo:2016} generates a weighted graph with synthesized classes that ensure the alignment between semantic embedding space and the classifier space by minimizing the distortion error.
BiDiLEL~\cite{wang:2017} learns two projection functions for projecting visual and semantic spaces onto a shared embedding space to preserve the relationship between them.

OutDist~\cite{mandal:2019} learns a visual feature synthesizer given the semantics and an out-of-distribution detector to distinguish generated features from seen ones.
\VE{WGAN~\cite{xian:2018wgan} is a model that synthesizes CNN features conditioned on class-level semantic information. It provides a way to generate a class-conditional feature distribution conditioned by a semantic descriptor.}
E2E~\cite{brattoli:2020} learns a CNN to generate visual features for unseen classes by training (in an end-to-end manner) this model with a combined dataset taking classes from Kinetics400 and overlapping classes of UCF101 and HMDB51.
Finally, CLASTER~\cite{gowda:2022claster} applies reinforcement learning on the clustering of visual-semantic~embeddings\footnote{\VE{A more detailed description for these methods can be found in~\cite{estevam:2021}}.}.

\subsection{Results}

\VE{Table~\ref{tab:likegowda2021} shows a comparison with the selected baselines. As can be seen, the proposed method achieves state-of-the-art performance on the UCF101 dataset, even without using the $67$ classes from the training set. The HMDB51 dataset is challenging due to their actions (e.g., run, turn, punch, chew, clap) are complex to define through text and due to their short video clips that do not take advantage of the Transformer architecture benefits. Despite these issues, we obtain a remarkable~performance.}

\input{tables/sota_truze}

\VE{\ac{zsar} has an extensive literature, with several strategies for performing video embedding and class embedding, as detailed in~\cite{estevam:2021}. Comparing these methods is not straightforward because several details on split configuration, random runs, and \ac{zsar} constraints must be taken into account. As mentioned previously, several deep learning-based video embeddings violate the ZSAR assumption when using 50\% of the classes for testing. Considering that several works fail in preserving this premise~\cite{lee:2021,sun:2022,piergiovanni:2020,mandal:2019}, a comparison under $50$\%/$50$\% or $0$\%/$50$\% protocols clarifies how good our method is compared to the broad literature.} 

\input{tables/sota_convetional}

\VE{Table~\ref{tab:sotalikexuetal} summarizes the performance on HMDB51 and UCF101 datasets for 28 different methods including ours. In this table, FV = fisher vector, BoW = bag of words, Obj = objects, S = image spatial feature, A = attribute, $W_{N}$ = word embedding of class names, $W_{T}$ = word embedding of class texts, ED = elaborative description, and Sent = sentences are the strategy adopted to perform video embedding. When the model uses a different number of classes in training, we indicate this by including this number next to the accuracy value.} 

\VE{There are two sections in Table~\ref{tab:sotalikexuetal}. The first groups the methods evaluated in the $50$\%/$50$\% protocol, whereas the second groups the methods evaluated in the $0$\%/$50$\% protocol (\ie \textit{cross-dataset}).}

\VE{To compare the results, we follow~\cite{estevam:2021} and assume that the mean accuracy has a normal distribution and approximate the population standard deviation $\sigma$ by sample standard deviation $s$. Therefore, the mean accuracy of population can be estimated by $\mu \approx \bar x \pm E$, where $E \approx t_{95\%, n-1} \frac{s}{\sqrt{n}}$ and $n-1$ are the degrees of freedom for $n$ runs.}

\VE{Considering this, we compare our results against the methods in which it is possible to estimate the mean accuracy with an error of $2\%$ at $95\%$ of confidence. Regarding the performance on UCF101, our method is on par with ER-ZSL, UR, SignleGAN, CLASTER (no statistical difference), which is impressive considering that it is based entirely on transfer learning. Methods such as E2E, PS-ZSAR or ViSET-96 are not directly comparable to our method since they do not provide the standard deviation value.} 

\VE{Finally, comparing our approach with methods that also use i3D for visual embedding, the proposed method is on par with CLASTER and outperforms GAN-KG, SFGAN, LMR, and OutDist by a large margin, demonstrating that its high performance is not only due to the bias from using~i3D. Unfortunately, we cannot quantify the underestimation performance due to disregarding the training split since HMDB51 and UCF101 datasets have no sentence~annotations.}

\VE{Considering the performance of our method on HMDB51 under 0/50\%, it is superior to O2A. It is worth mentioning that this dataset was not used in the evaluation of other methods in this group, possibly because it is very challenging to overcome the semantic gap due the simple actions. As an example, ER-ZSL~\cite{chen:2021} leverages object semantics in this dataset, but it improves generalization by concatenating visual features, which seems imperative to achieve higher performances as those obtained by CLASTER or SPOT.}

\subsection{Ablation Studies}

Here, we present a set of questions and answers \textit{Q\&A} to demonstrate the effectiveness of our approach. In all experiments, we use the same observers from the results shown in Table~\ref{tab:likegowda2021}.

\subsubsection{\VE{What is the impact of each observer or combination of observers on the performance?}}

In Table~\ref{tab:observersaccuracy}, we show the \ac{zsar} performance considering each observer individually, as well as some combinations of them. There is a huge difference in the accuracy rates achieved in the HMDB51 and UCF101 datasets, taking the same captioning models. Therefore, we discuss the results for each dataset~separately.

\input{tables/observers_acc_truze_ucf_hmdb}

In the UCF101 dataset, we observe that combining multiple observers has a considerable impact on performance.
The complete model is $27$\% (\ie~$49.1/38.6$) more accurate than the best observer individually. 
This property is a clear advantage of our model since new observers can be included later, thus improving overall performance. Another interesting case is the inclusion of OB2, which uses i3D and VGGish (see Fig.~\ref{fig:featuresandobservers}(b)). As mentioned earlier, approximately $50$\% of the videos have audio signal. However, this observer has a high individual performance and increases the final result by $2.3$\% (\ie$49.1/48$) compared to the best performance without~it.

Regarding the HMDB51 dataset, we believe that it is a challenging dataset for our approach mainly due to the short length of the videos (\ie~just $3.2$ seconds on average), which implies short stacks of features that nullify the benefits from self and multi-modal attention mechanisms.
This is evidenced by the fact that observers with different inputs do not learn better descriptions, as with the UCF101 dataset. 

In order to investigate \VE{the impact of stack length}, we extract features by reducing the frame stack length to 10 and 16 frames, corresponding to one i3D feature at 0.40 and 0.64 seconds, respectively. Table~\ref{tab:hmdb51} shows the results acquired with these features taking the same pre-trained models used in Table~\ref{tab:observersaccuracy}.
Notably, the performance is improved by $38$\%, considering the best cases from both tables ($20.4/14.8$). 

We note that, for this particular dataset, it is better to consider only observers based on Transformer models. This can be explained based on the characteristics of Visual GloVe features, which encode co-occurrence of visual patterns in complex events with long duration (one minute on average with a window of $24$s)~\cite{estevam:2021b}.
Hence, \ac{bmt}-based observers are not suitable for this dataset.
On the other hand, Visual GloVe proves to be useful as a feature enricher with Transformer (observer~OB3), as evidenced by the increase of $7$\% (OB1+OB3) compared to the~\ac{i3d} version alone (observer~OB1)~(\ie$20.4/19.1$).

\input{tables/observers_acc_hmdb_10_16frames}

\subsubsection{Is human involvement necessary for action class representation?}

Chen and Huang~\cite{chen:2021} introduced a method based on \ac{ed} (\ie a concatenation of class name and its sentence-based definition).
These descriptions were constructed by crawling candidate sentences from Wikipedia and dictionaries using action names as queries.
Afterward, annotators were asked to select and modify a minimum set of sentences. Table~\ref{tab:elabdescriptionsvsours} compares the \ac{zsar} performance considering four scenarios: only class label, \ac{ed}, Ours + \ac{ed}, and only~Ours. 

The results in both datasets show that the proposed pre-processing method achieves a higher accuracy compared to others.
Although~\ac{ed} reached impressive results in~\cite{chen:2021}, it did not prove efficient for adoption with our method, in which the joint embedding (visual and semantic) is based exclusively on transfer learning from the \ac{nlp} domain.
We believe this occurs due to the lack of fine-tuning with the descriptions of training classes in our~method.

\begin{table}[!htb]
\centering
\caption{\VE{ZSAR performance on the HMDB51 and UCF101 datasets under TruZe protocol considering different semantic information modalities.}}
\vspace{0.5mm}
\resizebox{.65\linewidth}{!}{
\begin{tabular}{lcc}
\toprule
                                           &  HMDB51        & UCF101        \\
\midrule
Baseline (only label)                      &  19.5          & 36.6          \\
Elaborative Descriptions~\cite{chen:2021} &  14.1          & 32.5          \\
Ours + Elaborative Descriptions            &  19.4          & 43.9          \\
\midrule
Ours                                       &  \textbf{20.4} & \textbf{49.1} \\
\bottomrule
\end{tabular}
}
\label{tab:elabdescriptionsvsours}
\end{table}

Considering these results, we propose the following question:

\subsubsection{How many sentences are required, and how is the ideal minimum length to represent class labels?}

\begin{figure}[!htb]
    \centering
    \captionsetup[subfigure]{captionskip=-0.25pt,font={scriptsize},justification=centering} 
    \subfloat[][]{
    \resizebox{0.492\linewidth}{!}{
    \includegraphics[height=9ex]{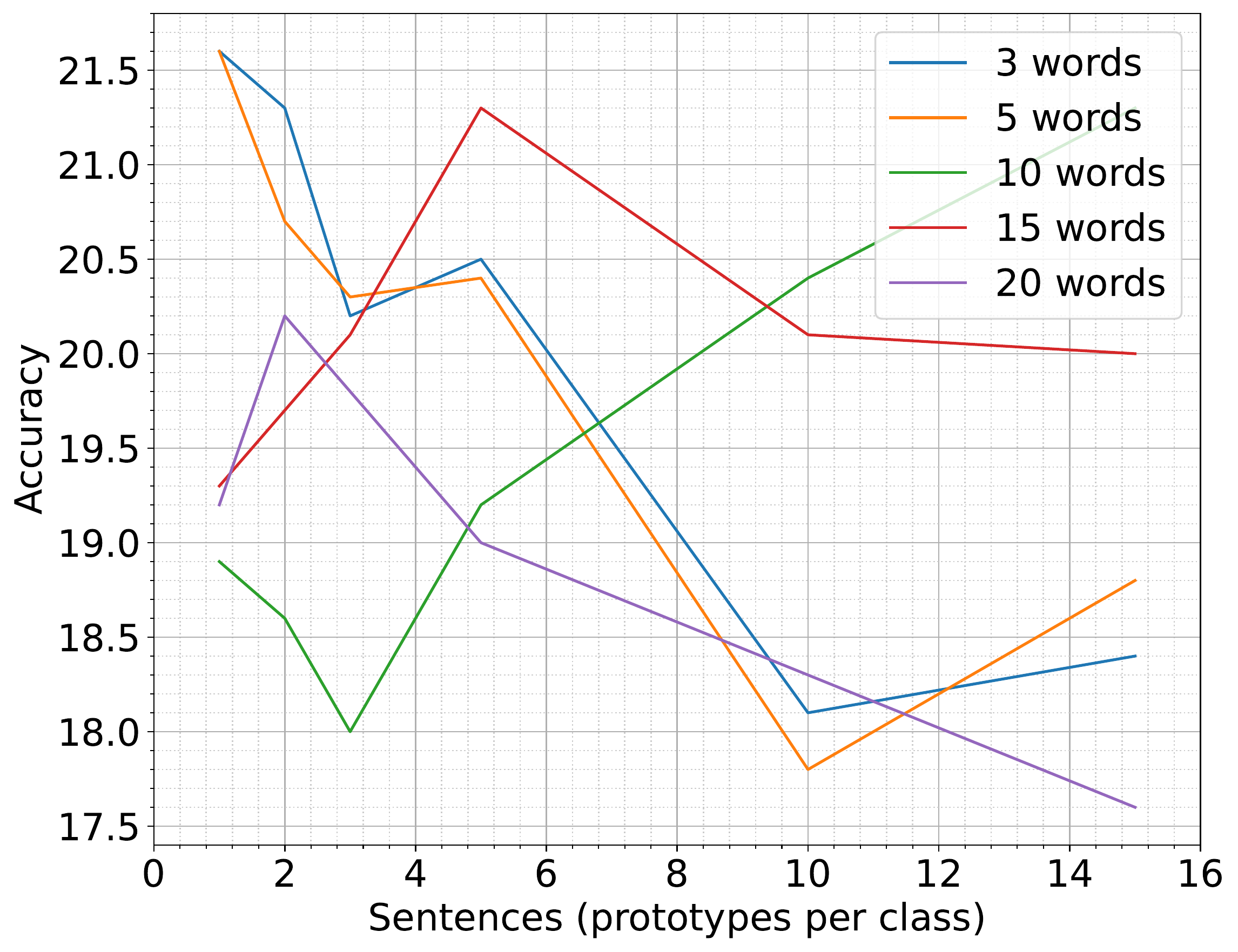}
    }}
    \subfloat[][]{
    \resizebox{0.487\linewidth}{!}{
    \includegraphics[height=9ex]{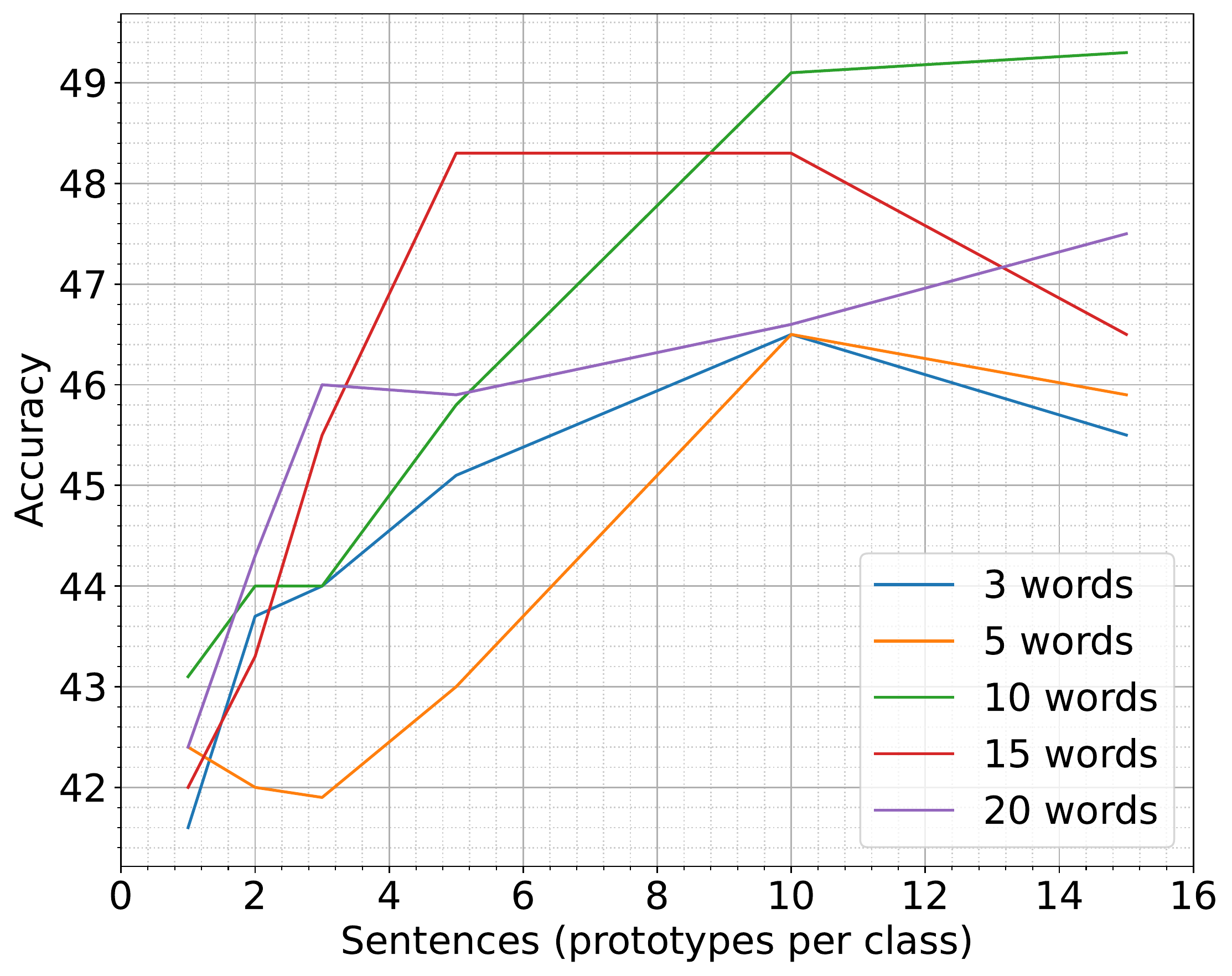}
    }
    }

    \caption{\VE{ZSAR performance changing the maximum sentences per class and the minimum words per sentence in the prototypes. (a) results from HMDB51 and (b) from UCF101.}}
    \label{fig:wordsandsentences}
\end{figure}

Fig.~\ref{fig:wordsandsentences}(a) and~\ref{fig:wordsandsentences}(b) show the accuracy considering a minimum length of $3$, $5$, $10$, $15$ and $20$ words per sentence for HMDB51 and UCF101, respectively.
We change the maximum number of sentences per class (\ie the number of prototypes in semantic space for each class) for each minimum length~value.

The graphs clearly show the need to balance the number of words and the number of sentences. There is a tendency for decreasing performance as more sentences are considered in HMDB51 and, conversely, an increasing in UCF101. Using short sentences, we inevitably select loose sentences containing the class label (\ie section titles or image labels in HTML pages), thus failing to capture the semantic context. On the other hand, when selecting long sentences with $15$ or $20$ words, we restrict the model to long explanations, failing to capture the immediate context of the class label. Therefore, our configuration (minimum of 10 words and up to 10 sentences) is a good trade-off between a minimum set of words and a maximum number of sentences in both~datasets. 

Additionally, the graph from Fig.~\ref{fig:wordsandsentences}(a) illustrates another aspect of why HMDB51 is so challenging for our method. The configurations with $3$ or $5$ words and only one sentence present the better performance, possibly because some actions in this dataset (\eg chew, pick, turn and wave) are semantically represented with a dictionary-style description (\ie short and precise descriptions).
This behavior is also evidenced in Table~\ref{tab:elabdescriptionsvsours}.

\subsubsection{Should we represent the class labels with separated sentences or with a paragraph?}

We can represent each class label with sentences or with a paragraph composed of the same sentences concatenated.
Table~\ref{tab:sentencesorparagraphs} shows the results taking only the class label (\ie one prototype per class, a single paragraph (\ie one prototype per class), or ten sentences (\ie ten prototypes per class).

Using sentences proves to be more accurate than the other options in both datasets. This characteristic is a remarkable aspect of our approach because other \ac{zsar} methods always consider only one prototype. 
Additionally, the paragraph representation proves to be better than the label name for our approach on UCF101. Indeed, the label name is insufficient for transferring knowledge from the language domain to the ZSAR classification.
Table~\ref{tab:sentencesorparagraphs} also suggests that the primary limitation on HMDB51 is related to the video sentence because there are no significant variations in accuracy taking different class label representations as there are on~UCF101.

\begin{table}[!htb]
\centering
\caption{\VE{Performance on the HMDB51 and UCF101 datasets under TruZe protocol considering separated sentences or paragraphs.}}

\vspace{0.5mm}

\resizebox{.65\linewidth}{!}{
\begin{tabular}{lcc}
\toprule
                                           &  HMDB51        & UCF101        \\
\midrule
Baseline (only label)                      &  19.5          & 36.6          \\
Paragraph                                  &  19.5          & 43.2          \\
Sentences                                  &  20.4          & 49.1          \\
\bottomrule
\end{tabular}
}
\label{tab:sentencesorparagraphs}
\end{table}

\subsubsection{How is the performance affected if we change the language encoder?}

Our method uses language encoders in two steps. In the first one, the encoder estimates the similarity between sentences from Internet documents and class labels, producing a semantic sentence space.
In the second step, the encoder embeds sentences from semantic space and video observers to generate a joint embedding space. 

We can employ different language encoders in these two steps, as shown in Table~\ref{tab:sentenceembedders}.
More specifically, we employ the Sentence2Vec~\cite{pagliardini:2017} model and two paraphrase models from the \textit{Sentence Transformers} repository: paraphrase-MiniLM-L6-v2 and paraphrase-distilroberta-base-v2. \VE{They are referred in Table~\ref{tab:sentenceembedders} as Sent2Vec, MiniLM, and DR, respectively}.
No models are fine-tuned or pre-trained with our data. The results clearly show that encoding the joint embedding space with Sentence2Vec is unsuitable since this model cannot overcome the gap between videos and class label descriptions, resulting in an accuracy close to the random~value.

On the other hand, the adoption of pre-trained paraphrase-based models results in a strong performance because the model is optimized to learn similarities in sentence pairs.
Using Sentence2Vec to pre-process the semantic information does not degrade the model performance at all. In this case, it is important to highlight that the comparison is made between the class label (which is not a sentence) and sentences. Therefore, this model can select sentences containing the exact label or synonyms. The performance combining Sentence2Vec with any paraphrase-based is lower than other configurations, possibly because the video descriptions are not enforced to present words contained in the class label in their~sentences.

\begin{table}[!htb]
\centering
\caption{\VE{Investigation on the semantic embedder for semantic pre-processing and \ac{zsar} embedding. Experiments performed on the TruZe protocol.}}
\vspace{0.5mm}
\resizebox{.80\linewidth}{!}{
\begin{tabular}{ccc|ccccc}
\toprule
\multicolumn{3}{c|}{Sem. Inf. Pre-proc.} & \multicolumn{3}{c}{ZSAR embedder} & \multirow{2}{*}{HMDB51} & \multirow{2}{*}{UCF101}\\
Sent2Vec & MiniLM   & DR      & Sent2Vec   & MiniLM   & DR     &          &         \\
\cmark   &          &         & \cmark     &          &        & \phantom{0}4.8      & \phantom{0}2.6     \\
\cmark   &          &         &            & \cmark   &        & 18.3     & 40.7    \\
\cmark   &          &         &            &          & \cmark & 16.0     & 40.4    \\
         & \cmark   &         & \cmark     &          &        & \phantom{0}7.5      & \phantom{0}1.5    \\
         & \cmark   &         &            & \cmark   &        & 19.9     & 45.9    \\
         & \cmark   &         &            &          & \cmark & 19.9     & 48.2    \\
         &          & \cmark  & \cmark     &          &        & \phantom{0}5.0      & \phantom{0}1.3     \\
         &          & \cmark  &            & \cmark   &        & 20.5     & 46.3    \\
         &          & \cmark  &            &          & \cmark & 20.4     & 49.1    \\
\bottomrule
\end{tabular}
}
\label{tab:sentenceembedders}
\end{table}

The observations in this experiment conduct us to the next question.

\subsubsection{What are the main limitations of our method?}

\begin{figure}[!htb]
    \centering
    \includegraphics[width=0.65\linewidth]{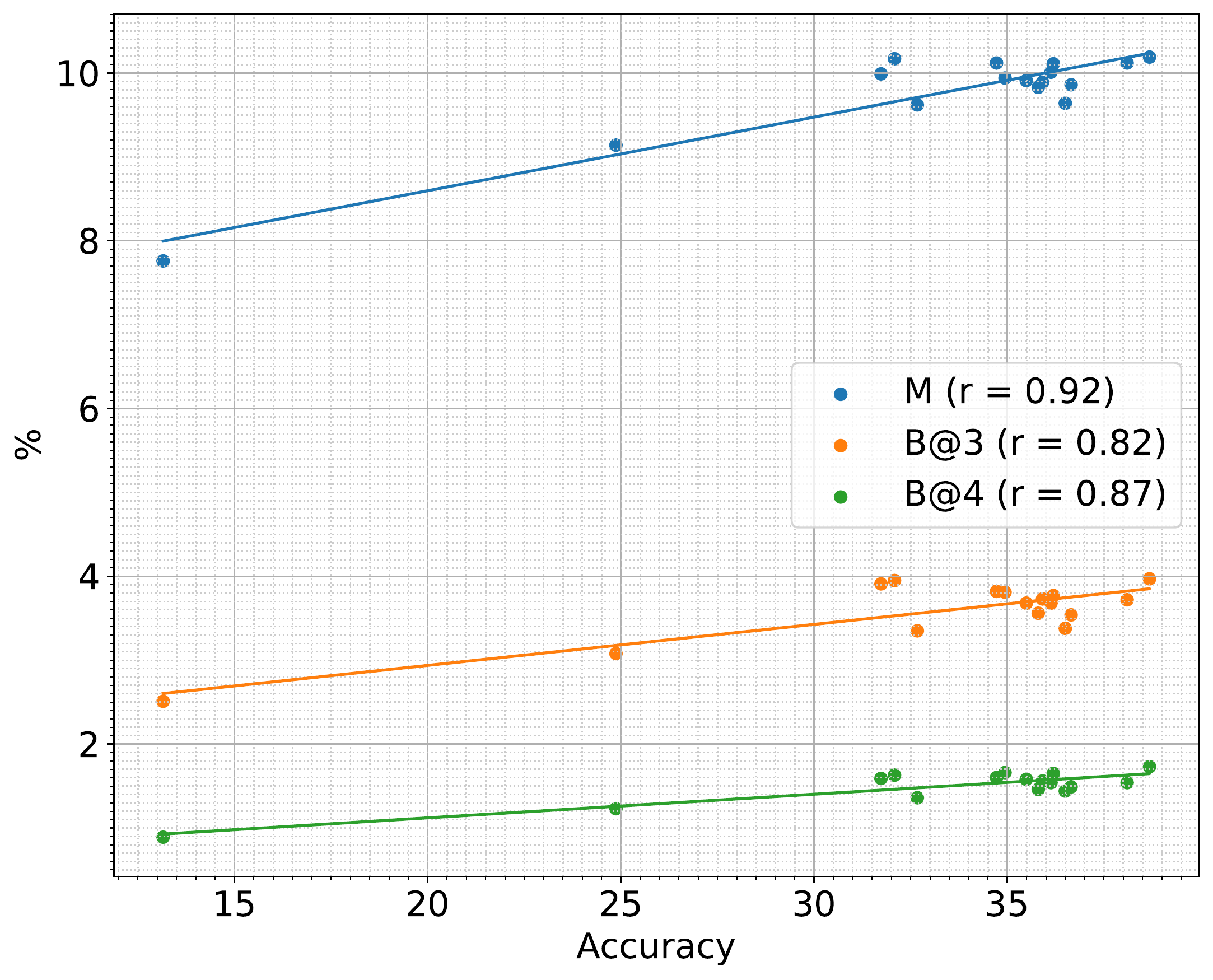}    
    \vspace{-1mm}
    \caption{Comparison of captioning scores (Meteor, Bleu 3, and Bleu 4) and ZSAR accuracy under the TruZe protocol for Observer 1 at different training stages.}
    \label{fig:correlation}
\end{figure}

\VE{In this subsection, we investigate two limiting aspects of our approach: the current \ac{sota} in video captioning and the inter-class similarity.
First, we examine the limitation of \ac{sota}} by taking the model from \textit{Observer 1} to compute the quality captioning measures (Meteor, Bleu@3, and Bleu@4) and \ac{zsar} accuracy for each training epoch on UCF101. The training was halted after ten epochs without improvements in Meteor.
As expected, there is a strong correlation ($r>0.8$) between these measures, especially on Meteor ($r>0.9$), as shown in Fig.~\ref{fig:correlation}.
Considering that video captioning is an active research topic with much room for improvement, the results suggest that better models for this task will directly lead to higher~accuracy.

\major{To conduct a more comprehensive investigation into inter-class performance, we selected a subset of 15 classes from UCF-101 that present challenging examples due to their high inter-class similarity. These classes can be divided into six groups: (1) activities involving horses, such as \textit{horse riding} and \textit{horse race}; (2) gymnastic performances, including \textit{pommel horse}, \textit{balance beam}, and \textit{floor gymnastics}; (3) activities involving basketballs, such as \textit{basketball} and \textit{basketball dunk}; (4) boxing-related actions, namely \textit{boxing punching bag} and \textit{boxing speed bag}; (5) activities involving the face, such as \textit{applying eye makeup}, \textit{applying lipstick}, and \textit{brushing teeth}; and (6) actions related to hair, such as \textit{blow drying hair}, \textit{getting a haircut}, and \textit{receiving a head~massage}.}

\begin{figure*}[!htb]
    \centering    
    \resizebox{0.80\linewidth}{!}{
    \includegraphics[height=20ex]{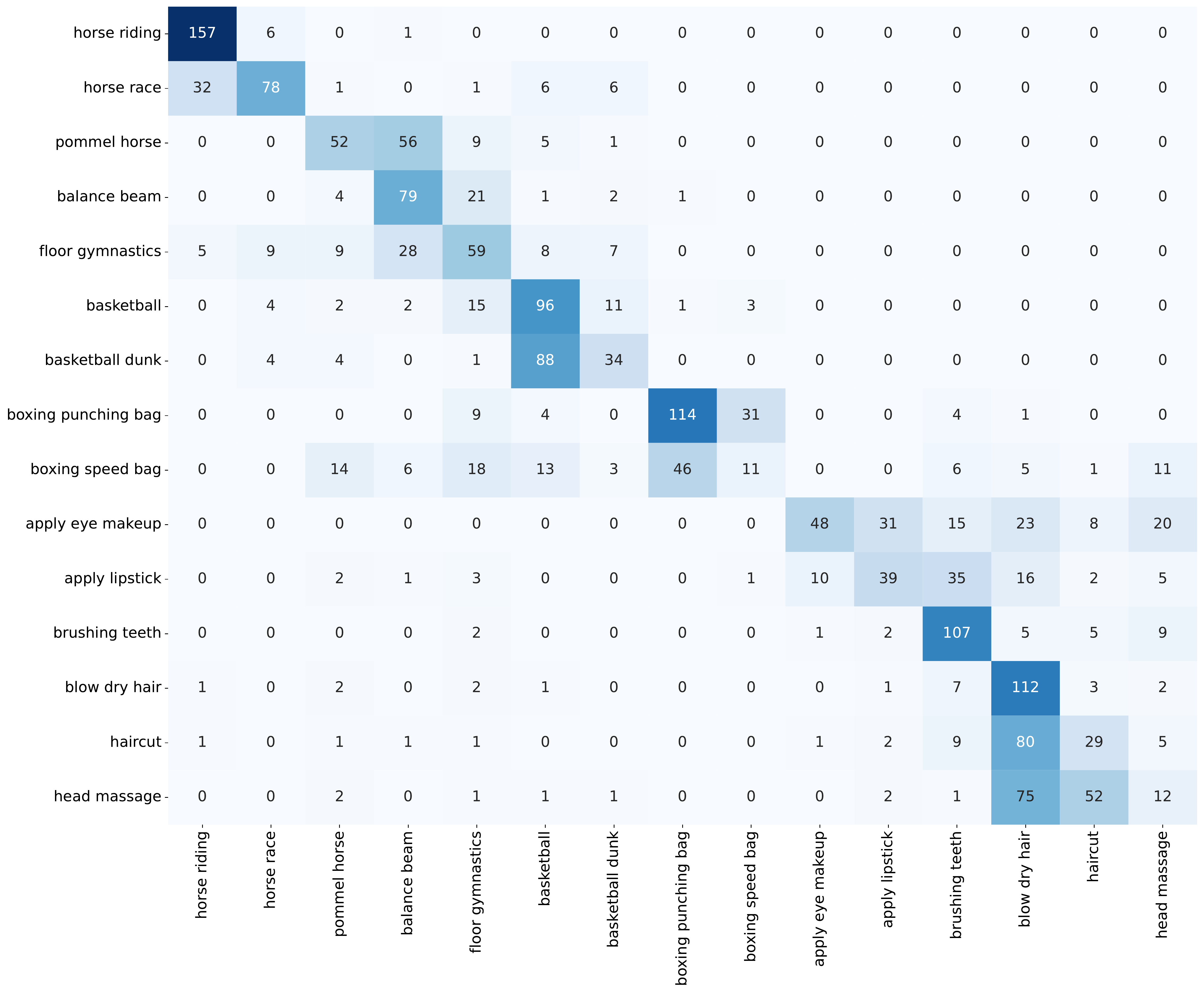}
    }        
    \vspace{0.5mm}
    \caption{\VE{Evaluation on the inter-class performance considering the complete method (5 observers) on UCF101.}}
    \label{fig:confmatrix}
\end{figure*}

\VE{Fig.~\ref{fig:confmatrix} clearly shows that the primary cause of errors lies in the high inter-class similarity (e.g., subgroups 4 -- boxing-related, 5 -- involving the face, 6 -- related to hair). The results indicate the need to extract more discriminative features from individual frames or short clips, which can be accomplished by incorporating object relationships or other semantic~features.}

\section{Conclusions and Future Work}
\label{sec:conclusions}

In this work, we proposed to perform \ac{zsar} by representing videos and semantic information with a common type of data: sentences in natural language.
We trained two video captioning architectures with different input modalities in the ActivityNet Captions dataset and used these models to produce sentences for the HMDB51 and UCF101 videos. We then evaluated the \ac{zsar} performance in a cross-dataset~scenario. Our conclusions are:

\begin{enumerate}
    \item \major{The textual descriptions provided by Observers proved to be sufficient for outperforming state-of-the-art performance on UCF101 and achieving remarkable results on HMDB51, even considering the relatively shorter time duration of clips in HMDB51 compared to UCF101. Nevertheless, it is necessary to consider a combination of Observers to achieve better~results;}
    \item \major{ZSAR can be effectively conducted using pre-trained paraphrase models, capitalizing on the abundance of available data, without requiring any additional training or domain adaptation techniques;}
    \item \major{We demonstrated a correlation between Meteor score and ZSAR accuracy, highlighting that the primary factor limiting performance is the current state of the art in video captioning. The proposed method is ``plug and play,'' allowing for the seamless replacement of models with more accurate ones as they become available. Furthermore, future research can explore the integration of captioning and ZSAR into an end-to-end model, optimizing their shared objectives;}
    \item \major{We specifically focused on working with captioning models in this study, but it is worth noting that models for various other tasks can also be employed to offer semantic information; for example, object detection with replacing by concepts (as in~\cite{chen:2021}) or video tagging.
    We acknowledge these possibilities and plan to investigate them in future~research.}
\end{enumerate}

\backmatter

\bmhead{Acknowledgments}
This work was supported by the Federal Institute of Paran\'{a}, Federal University of Paran\'{a} and by grants from the National Council for Scientific and Technological Development (CNPq) (grant numbers 304836/2022-2 and 308879/2020-1). The Titan Xp GPU used for this research were donated by the NVIDIA Corporation.

\section*{Declarations}
The authors declare that they have no known competing financial interests or personal relationships that could have appeared to influence the work reported in this paper. Our code is available at \url{https://github.com/valterlej/zsarcap}.

\bibliography{short,references}

\end{document}

%% file: 0-packages_commands.tex
\usepackage[nolist,nohyperlinks]{acronym}
\usepackage[caption=false,farskip=0pt]{subfig}
\usepackage{pifont}
\usepackage[numbers]{natbib}
\newcommand\VE[1]{{\textcolor{black}{#1}}}
\newcommand\major[1]{{\textcolor{black}{#1}}}

\newcommand{\ie}{{i.e.,~}}
\newcommand{\eg}{e.g.,~}
\newcommand{\etal}{{\em et al.}}
\newcommand{\cmark}{\ding{51}}

\DeclareMathOperator*{\argmax}{arg\,max}
\newcommand\copyrighttext{%
  \scriptsize Published at Multimedia Tools And Applications (\href{https://doi.org/10.1007/s11042-023-16566-5}{10.1007/s11042-023-16566-5}).}
\jyear{2022}
\theoremstyle{thmstyleone}

\theoremstyle{thmstyletwo}

\theoremstyle{thmstylethree}

%% file: 1-acronyms.tex
\begin{acronym}
\acro{bisst}[Bi-SST]{Bidirectional Single-Stream}
\acro{bert}[BERT]{Bidirectional Encoder Representations from Transformers}
\acro{bmt}[BMT]{Bi-Modal Transformer}
\acro{c3d}[C3D]{Convolutional 3D Network}
\acro{cnn}[CNN]{Convolutional Neural Network}
\acro{dap}[DAP]{Deep Action Proposal}
\acro{ed}[ED]{Elaborative Descriptions}
\acro{fps}[FPS]{frames per second}
\acro{gcn}[GCN]{Graph Convolutional Network}
\acro{glove}[GloVE]{Global Vectors}
\acro{visglove}[VisGloVE]{Visual Global Vectors}
\acro{gru}[GRU]{Gated Recurrent Unit}
\acro{har}[HAR]{Human Action Recognition}
\acro{i3d}[I3D]{Inflated 3D Network}
\acro{idt}[IDT]{Improved Dense Trajectories}
\acro{lstm}[LSTM]{Long Short-Term Memory}
\acro{mdvc}[MDVC]{Multi-Modal Dense Video Captioning}
\acro{mse}[MSE]{Mean Squared Error}
\acro{nlp}[NLP]{Natural Language Processing}
\acro{nn}[NN]{Nearest Neighbour}
\acro{of}[OF]{Optical Flow}
\acro{rnn}[RNN]{Recurrent Neural Network}
\acro{sbert}[SBERT]{Sentence-BERT}
\acro{sota}[SOTA]{state-of-the-art}
\acro{svm}[SVM]{Support Vector Machine}
\acro{tac}[TAC]{Trimmed Action Classification}
\acro{zsar}[ZSAR]{Zero-Shot Action Recognition}
\acro{zsl}[ZSL]{Zero-Shot Learning}
\end{acronym}

%% file: tables/tableofsymbols.tex
\begin{table*}[!htb]
\centering
\caption{\VE{Nomenclature used in our work.}}
\vspace{0.5mm}
\resizebox{.85\linewidth}{!}{
\begin{tabular}{cl}
\toprule
Notation                                           & Description                                             \\
\toprule
$\mathcal{Y}_{s}$, $\mathcal{Y}_{u}$               & sets of labels for labeled and unlabeled action classes \\
$y_{u_s}$, $y_{u_n}$                               & seen and unseen action classes                            \\
$y_{\text{pred}}$, $y_{{\text{prot}}}$             & predicted class label, prototype of an action class (a descriptive sentence)                                           \\
$\mathcal{Y}_{u_{\text{prots}}}$                   & set of label representations (textual description) for unlabeled action classes \\
$\emptyset$                                        & empty set                                               \\
$v$                                                & a video (visual and audio streams)                      \\
$n_c$                                              & a stack of video features for a video    \\
$v_f$                                              & a set of $n_c$ visual features           \\
$Y$                                                & a set of $m$ words of a sentence encoded with BMT or Transformer  \\
$d_{\text{model}}$                                 & dimension of the internal encoding layer of BMT or Transformer   \\
$Q$, $K$, $V$                                      & queries, keys and values (inputs of a self-attention layer)    \\
$d_k$                                              & dimension of the queries and keys  \\
$W$, $W_1$, $W2$                                   & internal weight matrices   \\
$V_f$, $A$, $Sm$                                   & a visual stream, an audio stream, a semantic stream (VisGloVe~\cite{estevam:2021b})        \\
$ASm$                                              & an audio or a semantic stream depending on the observer input \\
$\text{pos}$                                       & the position of a feature or word in a BMT or Transformer input \\ 
$i$                                                & number of column indices used on positional encoding \\
\text{PE}                                          & indicates that the feature was yielded by a positional encoding layer \\
\text{self}                                        & indicates that the feature was yielded by a self-attention layer  \\
\text{-att}                                         & indicates that the feature was yielded by a multi-head attention layer \\
\text{FFN}                                         & indicates that the features were yielded by a feed-forward network \\
$u_a$, $u_b$                                       & sentences a and b feed to the Siamese BERT networks~\cite{reimers:2019} \\
$n_s$                                              & dimension of sentence embeddings  \\
$k$                                                & number of labels in the paraphrasing classification pre-training \\
$s_a$, $s_p$, $s_n$                                & sentence embeddings for anchors, positive, and negative sentences (Sentence-BERT training) \\
\bottomrule
\end{tabular}
}
\label{tab:tableofsymbols}
\end{table*}

%% file: tables/sota_truze.tex
\begin{table}[!htb]
\centering
\caption{SOTA comparison under the TruZe protocol~\cite{gowda:2021b}. tr/te = train/test split configuration; Acc = accuracy.}

\vspace{0.5mm}

\resizebox{.70\linewidth}{!}{
\begin{tabular}{lcccc}
\toprule
                                                    &  \multicolumn{2}{c}{HMDB51}    & \multicolumn{2}{c}{UCF101} \\
                                                    &  tr/te         & Acc.          & tr/te         & Acc.       \\ 
\midrule
Latem~\cite{xian:2016}~\tiny{(CVPR'16)}           & 29/22          & \phantom{0}9.4           & 67/34         & 15.9       \\
SYNC~\cite{changpinyo:2016}~\tiny{(CVPR'16)}      & 29/22          & 11.6          & 67/34         & 15.0       \\
BiDiLEL~\cite{wang:2017}~\tiny{(IJCV'17)}         & 29/22          & 10.5          & 67/34         & 16.0       \\
WGAN~\cite{xian:2018wgan}~\tiny{(CVPR'18)}        & 29/22          & 21.1          & 67/34         & 22.5       \\
OutDist~\cite{mandal:2019}~\tiny{(CVPR'19)}       & 29/22          & 21.7          & 67/34         & 23.4       \\
E2E~\cite{brattoli:2020}~\tiny{(CVPR'20)}         & 29/22          & 31.5          & 67/34         & 45.2       \\
CLASTER~\cite{gowda:2022claster}~\tiny{(ECCV'22)} & 29/22          & \textbf{33.2} & 67/34         & 45.8       \\ 
SPOT~\cite{gowda:2023}~\tiny{(CVPRW'23)}          & 29/22          & 24.0          & 67/34         & 25.5       \\
\midrule
Ours                                                & 0/22           & 20.4          & 0/34          & \textbf{49.1} \\
\bottomrule
\end{tabular}
}
\label{tab:likegowda2021}
\end{table}

%% file: tables/sota_convetional.tex
\begin{table}[!htb]
\centering
\caption{\VE{SOTA comparison under $50$\%~/~$50$\% and $0$\%~/~$50$\% splits reporting Top-1 accuracy (\%) $\pm$ standard deviation. Our results were computed with $50$ random runs.}}
\resizebox{.98\linewidth}{!}{
\begin{tabular}{lcccc}
\toprule
Method                                          & Video      & Class    & HMDB51              & UCF101       \\ 
\midrule
$\textbf{50\% / 50\%}$                      & & & & \\
DAP~\cite{lampert:2009}~\tiny{(CVPR'09)}      & FV         & A        & N/A                 & 15.9$\pm$1.2 \\
IAP~\cite{lampert:2009}~\tiny{(CVPR'09)}      & FV         & A        & N/A                 & 16.7$\pm$1.1 \\
HAA~\cite{liu:2011}~\tiny{(CVPR'11)}          & FV         & A        & N/A                 & 14.9$\pm$0.8 \\
SVE~\cite{xu:2015}~\tiny{(ICIP'15)}           & BoW        & $W_N$    & 13.0$\pm$2.7        & 10.9$\pm$1.5 \\
ESZSL~\cite{paredes:2015}~\tiny{(ICML'15)}    & FV         & $W_N$    & 18.5$\pm$2.0        & 15.0$\pm$1.3 \\
SJE~\cite{akata:2015}~\tiny{(CVPR'15)}        & FV         & $W_N$    & 13.3$\pm$2.4        & \phantom{0}9.9$\pm$1.4  \\
SJE~\cite{akata:2015}~\tiny{(CVPR'15)}        & FV         & A        & N/A                 & 12.0$\pm$1.2 \\
MTE~\cite{xuhospedales:2016}~\tiny{(ECCV'16)} & FV         & $W_N$    & 19.7$\pm$1.6        & 15.8$\pm$1.3 \\
ZSECOC~\cite{qin:2017}~\tiny{(CVPR'17)}       & FV         & $W_N$    & 22.6$\pm$1.2        & 15.1$\pm$1.7 \\
ASR~\cite{wang:2017b}~\tiny{(ECML PKDD'17)}   & C3D        & $W_T$    & 21.8$\pm$0.9        & 24.4$\pm$1.0 \\
UR~\cite{zhu:2018}~\tiny{(CVPR'18)}           & FV         & $W_N$    & N/A                 & 42.5~$\pm$~0.9~\tiny{(200)} \\
OutDist~\cite{mandal:2019}~\tiny{(CVPR'19)}   & i3D+C3D    & A        & N/A                 & 38.3$\pm$3.0 \\
OutDist~\cite{mandal:2019}~\tiny{(CVPR'19)}   & i3D+C3D    & $W_N$    & 30.2$\pm$2.7        & 26.9$\pm$2.8  \\
TS-GCN~\cite{gao:2019}~\tiny{(AAAI'19)}       & Obj        & $W_N$    & 23.2$\pm$3.0        & 34.2$\pm$3.1 \\
LMR~\cite{piergiovanni:2020}~\tiny{(WACV'20)} & i3D        & $W_N$    & 34.7$\pm$2.4        & 33.4$\pm$1.8 \\
E2E~\cite{brattoli:2020}~\tiny{(CVPR'20)}     & r(2+1)d    & $W_N$    & 32.7~\tiny{(664)}                & 48~\tiny{(664)}           \\      
SFGAN~\cite{lee:2021}~\tiny{(Neurocomputing'21)} & i3D     & $W_N$    & 32.4$\pm$4.1        & 29.8$\pm$2.8 \\
DASZL~\cite{kim:2021}~\tiny{(AAAI'21)}        & TSM        & A        & N/A                 & 48.9$\pm$5.8 \\
ER-ZSL~\cite{chen:2021}~\tiny{(ICCV'21)}      & (S+Obj)    & ED       & 35.3$\pm$4.6        & 51.8$\pm$2.9 \\
PS-ZSAR~\cite{kerrigan:2021}~\tiny{(NeurIPS 21)} & r(2+1)d   & $W_T$    & 33.8~\tiny{(664)}                & 49.2~\tiny{(664)} \\
GAN-KG~\cite{sun:2022}~\tiny{(PR'22)}         & i3D        & $W_N$    & 31.2$\pm$1.7        & 28.3$\pm$1.8 \\ 
Single-GAN~\cite{huang:2022singlegan}~\tiny{(VISAPP'22)} & i3D & ResNet101 & N/A           & 45.9$\pm$3.42 \\ 
CLASTER~\cite{gowda:2022claster}~\tiny{(ECCV'22)} & i3D    & $W_N$    & 41.8$\pm$2.1        & 50.2$\pm$3.8 \\
SPOT~\cite{gowda:2023}~\tiny{(CVPRW'23)}      & i3D+C3D + SPOT & $W_N$ & 39.8$\pm$1.4        & 42.8$\pm$1.7 \\
ViSET-96~\cite{doshi:2022}~\tiny{(CVPRW'23)}  & ViSET      &  $W_T$   & 34.5~\tiny{(564)}     &  \textbf{53.2}~\tiny{(564)} \\

\midrule \midrule \\[-2.75ex]
$\textbf{0\% / 50\%}$                       & & & & \\
O2A~\cite{jain:2015}~\tiny{(ICCV'15)}         & Obj        & $W_N$    & 15.6                & 30.3         \\
SAOE~\cite{mettes:2017}~\tiny{(ICCV'17)}      & Obj        & $W_N$    & N/A                 & 40.4$\pm$1.0 \\
% object priors
OP~\cite{mettes:2021}~\tiny{(IJCV'21)}        & Obj        & $W_N$    & N/A                 & 47.3  \\ % fasttext
DO-SC~\cite{bretti:2021}~\tiny{(BMVC'21)}     & Obj        & $S_{embs}$ & N/A               & 45.2$\pm$4.6 \\% MINI_LM sentence_transformers
Ours                                            & Sent    & Sent     & 28.3$\pm$3.0        & \textbf{49.0$\pm$3.5} \\
\bottomrule
\end{tabular}
}
\label{tab:sotalikexuetal}
\end{table}

%% file: tables/observers_acc_truze_ucf_hmdb.tex
% OB1 - transformer (i3d)
% OB2 - bmt (i3d) + (vggish)
% OB3 - transformer (i3d+visglove)
% OB4 - bmt (i3d) + (visglove)
% OB5 - bmt (i3d+visglove) + (visglove)
\begin{table}[!htb]
\centering
\caption{\VE{Observer accuracy for the UCF101 and HMDB51 datasets under TruZe protocol. No training classes were used to train the models.}}
\vspace{0.5mm}
\resizebox{.65\linewidth}{!}{
\begin{tabular}{ccccccccc}
\toprule
OB1    & OB2    & OB3    & OB4    &  OB5    &  HMDB51  & UCF101\\
\midrule
\cmark &        &        &        &         & 14.4     & 38.6 \\
       & \cmark &        &        &         & $-$      & 37.2 \\
       &        & \cmark &        &         & 13.5     & 34.6 \\
       &        &        & \cmark &         & 12.7     & 30.9 \\
       &        &        &        &  \cmark & 10.6     & 35.3 \\ \midrule
\cmark &        & \cmark &        &         & 14.8     & 44.9 \\
\cmark &        & \cmark & \cmark &         & 14.2     & 47.3 \\
\cmark &        & \cmark & \cmark & \cmark  & 14.5     & 48.0 \\ \midrule
% using audio signal
\cmark & \cmark &        &        &         & $-$      & 46.5 \\
\cmark & \cmark & \cmark &        &         & $-$      & 48.9 \\
\cmark & \cmark & \cmark & \cmark &         & $-$      & 48.9 \\
\cmark & \cmark & \cmark & \cmark &  \cmark & $-$      & \textbf{49.1} \\
\bottomrule
\end{tabular}
}
\label{tab:observersaccuracy}
\end{table}

%% file: tables/observers_acc_hmdb_10_16frames.tex
\begin{table}[!htb]
\centering
\vspace{0.5mm}
\resizebox{.65\linewidth}{!}{
\begin{tabular}{ccccccc}
\toprule
10     & 16     & OB1    & OB3    & OB4    &  OB5    &  HMDB51  \\ \midrule
\cmark &        & \cmark &        &        &         & 19.1     \\
\cmark &        &        & \cmark &        &         & 17.8     \\
\cmark &        & \cmark & \cmark &        &         & \textbf{20.4}     \\
\cmark &        &        &        & \cmark &         & 14.9     \\
\cmark &        &        &        &        & \cmark  & 14.3     \\
\cmark &        & \cmark & \cmark & \cmark & \cmark  & 19.1     \\ 
\midrule       
       & \cmark & \cmark &        &        &         & 19.2     \\
       & \cmark &        & \cmark &        &         & 16.6     \\
       & \cmark & \cmark & \cmark &        &         & 19.2     \\       
       & \cmark &        &        & \cmark &         & 16.5     \\
       & \cmark &        &        &        & \cmark  & 15.7     \\
       & \cmark & \cmark & \cmark & \cmark & \cmark  & 19.1     \\
\bottomrule
\end{tabular}
}
\label{tab:hmdb51}
\end{table}